\documentclass[preprint]{article}
\PassOptionsToPackage{sort&compress,numbers,super}{natbib}
\usepackage{neurips_2024}
\usepackage{multirow}%
\usepackage{amsmath,amssymb,amsfonts}%
\usepackage{amsthm}%
\usepackage{mathrsfs}%
\usepackage[title]{appendix}%
\usepackage{xcolor}%
\usepackage{textcomp}%
\usepackage{manyfoot}%
\usepackage{booktabs}%
\usepackage{algorithm}%
\usepackage{algorithmicx}%
\usepackage{algpseudocode}%
\usepackage{adjustbox}
\usepackage{listings}%
\usepackage[ruled, vlined, linesnumbered,algo2e]{algorithm2e}
\usepackage{wrapfig}
\usepackage[utf8]{inputenc} % allow utf-8 input
\usepackage[T1]{fontenc}    % use 8-bit T1 fonts
\usepackage{hyperref}       % hyperlinks
\usepackage{url}            % simple URL typesetting
\usepackage{booktabs}       % professional-quality tables
\usepackage{amsfonts}       % blackboard math symbols
\usepackage{nicefrac}       % compact symbols for 1/2, etc.
\usepackage{microtype}      % microtypography
\usepackage{xcolor}         % colors
\usepackage{graphicx}%
\theoremstyle{thmstyleone}%
\theoremstyle{thmstyletwo}%

\theoremstyle{thmstylethree}%
\DeclareMathOperator*{\argmax}{argmax}
\usepackage{tikz}
\usetikzlibrary{shapes.geometric, arrows, positioning}
\tikzstyle{decision} = [diamond, draw, fill=orange!30, 
    text width=5em, text centered, node distance=3cm, inner sep=0pt]
\tikzstyle{block} = [rectangle, draw, fill=gray!10, 
    text width=10em, text centered, minimum height=3em]
\tikzstyle{line} = [draw, -latex', thick]
\tikzstyle{yes} = [fill=green!20, text centered, node distance=2cm]
\tikzstyle{no} = [fill=red!20, text centered, node distance=2cm]
\tikzstyle{endblock1} = [rectangle, draw, fill=red!30, 
    text width=6em, text centered, minimum height=2em]
\tikzstyle{endblock2} = [rectangle, draw, fill=green!30, 
    text width=6em, text centered, minimum height=2em]

\usepackage{xcolor}
\usepackage{soul} % For highlighting

\begin{document}
\author{%
  Víctor Sabanza-Gil \\
  EPFL\\
  Switzerland \\
  \And
  Daniel Pacheco Gutiérrez\\
  Atinary Technologies Inc.\\
  Switzerland
  \And
  Jeremy S. Luterbacher\\
  EPFL\\
  Switzerland
  \And
  Riccardo Barbano \\ 
  Atinary Technologies Inc.\\
  Switzerland
  \And
  José M. Hernández-Lobato \\ 
  University of Cambridge \\
  United Kingdom \\
  \And
  Philippe Schwaller \\
  EPFL \\
  Switzerland\\
  philippe.schwaller@epfl.ch
  \And
  Loïc Roch\\
  Atinary Technologies Inc.\\
  Switzerland\\
  loic.roch@atinary.com
}

\title{Best Practices for Multi-Fidelity Bayesian Optimization in Materials and Molecular Research}

\maketitle

\begin{abstract}
{%
Multi-fidelity Bayesian Optimization (MFBO) is a promising framework to speed up materials and molecular discovery as sources of information of different accuracies are at hand at increasing cost.
Despite its potential use in chemical tasks, there is a lack of systematic evaluation of the many parameters playing a role in MFBO.
In this work, we provide guidelines and recommendations to decide when to use MFBO in experimental settings. 
We investigate MFBO methods applied to molecules and materials problems.
First, we test two different families of acquisition functions in two synthetic problems and study the effect of the informativeness and cost of the approximate function.
We use our implementation and guidelines to benchmark three real discovery problems and compare them against their single-fidelity counterparts.
Our results may help guide future efforts to implement MFBO as a routine tool in the chemical sciences.
}

\end{abstract}

\section*{Introduction}
Among Machine Learning (ML) techniques, Bayesian optimization (BO) has emerged as the go-to choice for optimizing the design of experiments in the chemical domain \cite{Guo2023-borxn, hase2018_phoenics, Shields2021-as, Braconi2023-bochem_rev}. 
Bayesian optimization, grounded in a probabilistic framework \cite{Garnett2023-iq}, consists of two main components: a probabilistic model that serves as a proxy for the experimental process being optimized, and a policy that governs the acquisition of new experimental data.
For instance, a researcher seeking to maximize the yield of a given reaction would query the model to identify which experimental conditions to be tested to achieve the desired outcome.
This methodology has shown success in diverse optimization tasks such as chemical reactions \cite{hase2018_phoenics,hase2018chimera,Guo2023-borxn},functional molecules \cite{Griffiths2022-photoswitches} or nanocrystal shape\cite{Zaza2025-uk}. Iterating through this learning cycle has recently enabled the rapid identification of optimal conditions within extensive search spaces \cite{Suvarna2024-zn, Li2024-xp}.

\begin{figure}[h]
\includegraphics[width=0.98\columnwidth]{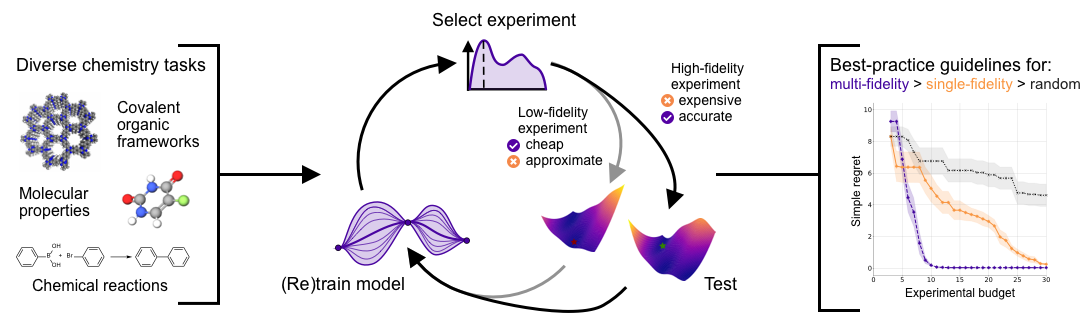}
\caption{Multi-fidelity Bayesian Optimization (MFBO) combines expensive but informative, and cheap but approximate sources of information with a ML model to optimize black-box problems at an overall reduced budget. The high-fidelity source (HF, e.g., real lab experiment) provides accurate information but at a high cost. The low-fidelity source (LF, e.g., computer simulation of real experiment) represents a cheaper approximation of the HF. In the illustration, we show the surfaces corresponding to the Branin function used in this study at the HF and one LF level, where the optima are located in different places. The MFBO loop iteratively selects HF or LF to maximize the information gain while reducing the overall optimization cost.}

\label{fig1}
\end{figure}

While canonical BO has recently been popularized among experimentalists, the experiment design may benefit if the practitioner can collect data at different degrees of reliability while paying a lower price.
Additional experimental evidence that may be readily available can be integrated within the model representing the process to be optimized (e.g., low-precision experiments conducted with bench-top nuclear magnetic resonance can be integrated into more expensive, high-precision experiments \cite{Blumich2019-benchtopnmr}). 
In classical experimental design, the inclusion of such information sources representing different reliabilities is referred to as multi-fidelity Bayesian Optimization (MFBO). 
Within this setting, a specific cost is assigned to each information source — hereafter defined as fidelity. 
The multi-fidelity probabilistic model learns the process of interest by extracting knowledge from data available at different fidelities and understanding their interplay. 
The policy for querying new experimental data at a given fidelity also takes into account the overall cost.
By combining low-fidelity (LF) and high-fidelity (HF) points, the overall optimization cost can be reduced compared to the single-fidelity BO (SFBO). 
Figure \ref{fig1} exemplifies how the iterative cycle of MFBO may reduce the overall cost, compared to the standard SFBO. 

Although MFBO has garnered interest in the past decade within the ML community, leading to a myriad of model definitions and acquisition policies \cite{bonilla_2007, NIPS2013_f33ba15e, wu2020, takeno20a}, researchers have only lately started to integrate them within their design methodologies. 
In the chemical domain, there has been recent interest in incorporating cost awareness into the BO loop\cite{Schoepfer2024-zu, costawareBO_2024_electr}. 
Regarding the specific multi-fidelity approach, several studies have successfully applied MFBO to the materials discovery domain \cite{Fare2022-yi, Gantzler2023-fj, Kim2024-multiBOWS, Folch2023-nn, Palizhati2022, Jacobs2023}, drug discovery\cite{macdonald2025_mfbo} or reactor design\cite{Savage2024-reactorsmfbo}.
However, there is a fundamental lack in the assessment of MFBO performance, and due to this, each work represents MFBO performance in different ways \cite{nips2016_MFUCB,Gantzler2023-fj, Palizhati2022, Jacobs2023}.
Although some metrics have been proposed \cite{Fare2022-yi, Palizhati2022}, the lack of clarity and unified criteria on how to assess the benefits of MFBO in the chemical domain has hindered its widespread adoption among the experimental community\cite{dovonon2023longrunbehaviourmfbo, judge2024}.
Several factors come into play when assessing its impact, yet the application of MFBO can also be detrimental to the overall optimization process \cite{mikkola23a}. 
It has been also shown how in the long run the advantage of MFBO over SFBO can be lost \cite{dovonon2023longrunbehaviourmfbo}. Therefore, it is crucial to unify and provide a reliable method to decide when MFBO is better than SFBO.

In this work, we propose a series of guidelines for when to use MFBO within the experimental design pipeline. 
We conduct an exhaustive experimental investigation on both standard MFBO problems — ubiquitous in BO literature — as well as chemistry-based ones.
Initially, we optimize two synthetic problems to assess the behavior of a multi-fidelity model in simulated black-box scenarios, detecting unfavorable situations where the application of MFBO does not offer an advantage over SFBO. 
We exhaustively scan MFBO experimental parameters (namely, cost ratio and informativeness of the LF source) to identify trends that indicate promising experimental settings for when to apply MFBO.
We then progress to chemistry-based problems, tackling three real challenges in molecular and materials optimization, where MFBO successfully outperforms SFBO. 
Finally, we extract guidelines and recommendations based on the synthetic and real benchmarks to inform prospective users on which scenarios can favor MFBO over SFBO.
This work offers a comprehensive view and a reference guide for applying MFBO in the molecules and materials discovery domain.

\section*{Results}
\subsection*{Shedding light into MFBO failure modes}\label{sec:failures}

\begin{figure}[h]
\centering
\includegraphics[width=0.95\columnwidth]{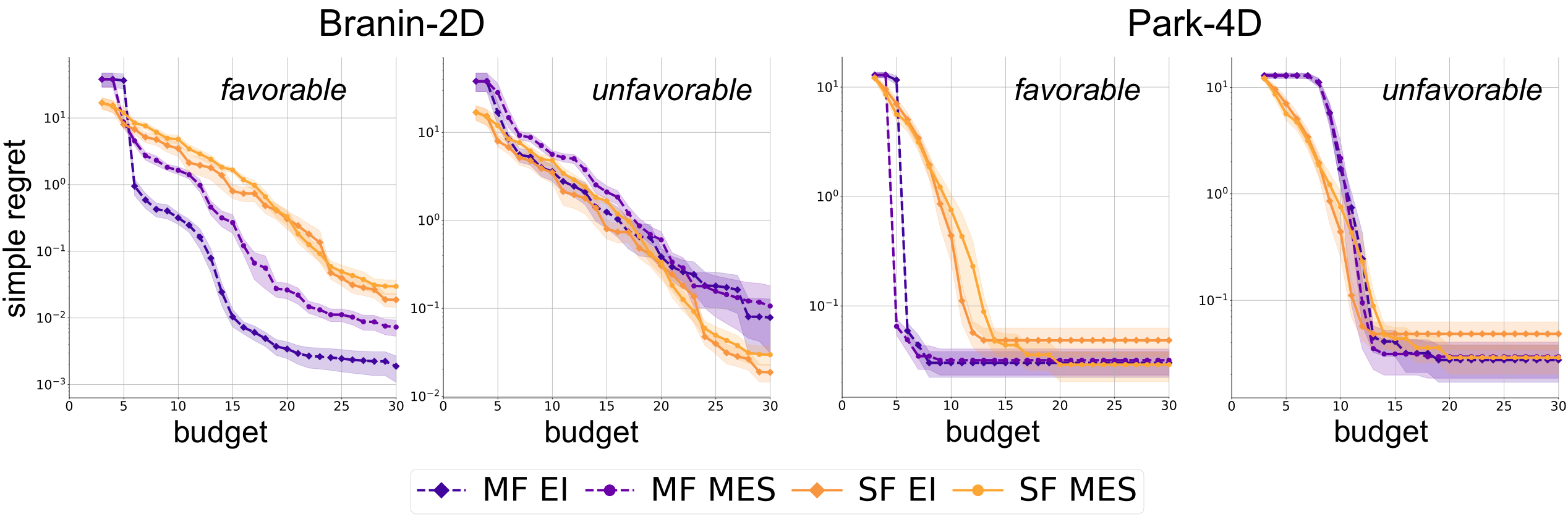}
\caption{MFBO results are dependent on problem conditions. In the favorable scenario (highly informative and cheap LF source, $\rho$ = 0.1, $R^2$ > 0.9), MFBO offers better performance than SFBO on the synthetic functions (maximum $\Delta$ of 0.53 and 0.33 for Branin and Park, respectively). The optimization traces show how lower regrets are found at lower budgets under this scenario. When the conditions are changed and a low informative and more expensive LF is used ($\rho$ = 0.5, $R^2$ < 0.75), MFBO performance is degraded. Under this scenario, the optimization traces are similar, and MFBO performance is notably reduced (minimum $\Delta$ of -0.07 and -0.11 for Branin and Park, respectively). The reference budget unit cost corresponds to a single HF evaluation.}
\label{fig3}
\end{figure}

We test the performance of MFBO methods over their SFBO counterparts on two different synthetic functions, Branin and Park (with 2 and 4 dimensions respectively), on two different scenarios (favorable and unfavorable). Synthetic functions are commonly employed as black-box problems to test the performance of BO algorithms (see \hyperref[synthetic_funcs]{Synthetic functions}). In the case of MFBO, the output of the functions can be biased using a parameter $\alpha$ to provide lower accuracy information sources. This parameter is related to the informativeness of the LF source (in our case, we quantify the informativeness of the LF by computing the $R^2$ with respect to the HF, see \hyperref[sec:metrics]{Metrics}). For the favorable scenario, we tune the $\alpha$ parameter to provide a highly informative LF source (LF $R^2 > 0.9$). We set the cost of querying a HF point to 1 and the cost of a LF point to 0.1 ($\rho$ = 0.1). Figure \ref{fig3} shows the results for the favorable scenario. In this case, both MFBO methods outperform their SFBO counterparts, with maximum discounts of 0.53 and 0.33 for Branin and Park respectively. The MFBO runs get to lower regrets quicker, spending less resources by exploiting the use of LF and HF sources of information. The resulting optimization runs are therefore desirable as they effectively leverage the access to the LF to guide the optimization of the HF level.

Although under the previous settings MFBO provides a higher performance than SFBO, a change in the LF conditions dramatically affects the result of the optimization. In the unfavorable scenario, we set the cost of the LF source to 0.5 (only half of the HF cost, $\rho$ = 0.5) and decrease the informativeness of the functions ($R^2$ $<$ 0.75). In this case, the performance of both methods decreases with respect to the favorable scenario. In Branin, the maximum $\Delta$ drops to -0.07, whereas in Park it drops to a maximum $\Delta$ of -0.11. In both cases the favorable trend is reverted, and MFBO loses its advantage over SFBO. These results illustrate how MFBO performance is affected by the cost and the informativeness of the LF source, losing its advantage over SFBO if the LF source is not informative and cheap enough. It also reflects how the MFBO setting is problem-dependent and it may be more robust towards problem changing situation 

Apart from problem conditions, model choice may also affect MFBO performance. Previous studies have proposed and used many MF models, mainly GPs refs but also DNNs\cite{NIPS2013_f33ba15e, NEURIPS2020_BO_DNN}. We compared the standard BoTorch multi-fidelity model with a MultiTask GP\cite{bonilla_2007} to see how model choice can affect the results. The main difference between the two models lies in how fidelities are encoded in the GP. SI section \ref{secA1} shows a preliminary study on the fidelity kernel and how both models offer similar performances in regression and MFBO tasks with the Branin function (figures \ref{fig:mf_mt_regression} and \ref{fig:mt_branin} respectively). In addition, using two independent Gaussian Processes (GPs) for high-fidelity (HF) and low-fidelity (LF) data would fail to model the correlation between them, as each GP would learn its own independent function without leveraging shared structure. The key advantage of a multi-fidelity GP is that it introduces a covariance structure between HF and LF data with the kernel $k_IS$. This correlation enables the model to transfer information across fidelities, leading to more data-efficient learning and better uncertainty quantification, especially when HF data is scarce. As an example, the regression task in the Supplementary Information \ref{catalysis_task} shows how the incorporation of lower fidelities improves the uncertainty estimation at the HF level. Therefore, we stick to the use of the default model provided in BOTorch, focusing on the external factors that can affect the experiments. 

\subsection*{Finding suitable scenarios for MFBO}
\begin{figure}[h]
\includegraphics[width=0.95\columnwidth]{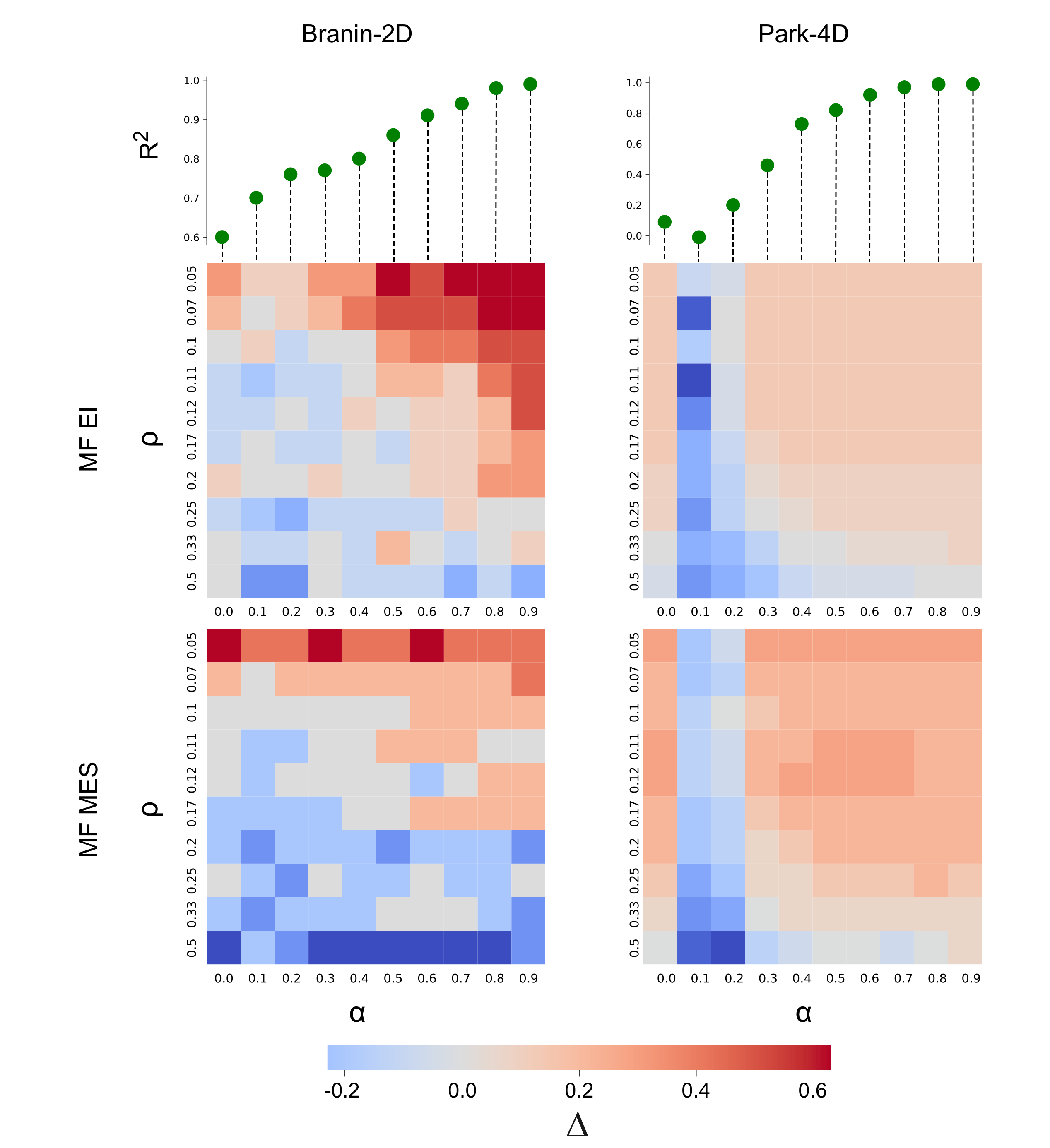}
\caption{Change of $\Delta$ with the cost and informativeness of the LF source. Each heatmap shows the computed $\Delta$ for each synthetic function and acquisition function family. The heatmap's $y$ axis represents the cost of the LF source ($\rho$), and the $x$ axis its informativeness. The informativeness of each function is tuned using a parameter $\alpha$ in the definition of the function. Because $\alpha$ cannot be known in a real experiment, the upper plots show the estimated $R^2$ value for each $\alpha$ value (see \hyperref[sec:costandinfo]{Methods} for an explanation of $R^2$ estimation). In all the cases, cheaper and informative LF sources provide the highest discounts (a gradient in $\Delta$ increasing when $\rho$ decreases and $R^2$ increases can be observed).}
\label{fig4}
\end{figure}

We investigate in detail the effect of the LF cost and informativeness on MFBO performance (measured by $\Delta$, see \hyperref[sec:metrics]{Metrics}) for the two previous synthetic functions. We run the MFBO method with different combinations of LF cost and informativeness and compare it to SFBO. Figure \ref{fig4} shows the computed $\Delta$ for each run as a result of the previous parameters. The $x$ axis of the heatmap represents the $\alpha$ value that was used to bias the synthetic function, and the upper plot shows the computed $R^2$ for each value. In both cases, a gradient can be observed where the progression towards cheaper and more informative LF sources provides better MFBO performances (higher $\Delta$). In the case of Branin, the lower values of $R^2$ in the synthetic functions are associated with a less informative LF approximation, which translates to lower values of $\Delta$. In the same way, in Park $\Delta$ is also decreased when the $R^2$ is lower, giving negative results when its value is close to 0. This result is consistent as a more inaccurate source of information is likely to reduce the performance of the MFBO method. In terms of cost, there is also an inverse correlation in both cases, where lower $\rho$ (cheaper LF sources) generate higher discounts. This is an expected trend, as the model can access the approximate sources at a lower cost, allowing a wider exploration of the problem space for the same price.

Although $\Delta$ is reported at a value of $\tau$ = 0.9 for all the cases, we also investigated the effect of $\tau$ on the final discount. In previous works, the MFBO performance was reported to be lower than SFBO at smaller budgets\cite{Jacobs2023}, and the authors noted that in the long-run MFBO may also lose its advantage over SFBO\cite{dovonon2023longrunbehaviourmfbo}. This trend was observed in both functions and AFs, where low values of $\tau$ and a $\tau$ of 1 provided negative discounts (see figure \ref{3dplot} in the SI). Although the temporal dependence of the MFBO is hard to control for the experimenter due to the uncertainty of when to stop the sampling, the previous figure shows how there exist "sweet spots" where MFBO can exploit the cheap LF source with the available budget to provide an advantage over SFBO.

The results follow the same trend in both cases independently of the acquisition function. Although there are differences in the absolute $\Delta$ values between the two problems (due to the specific landscape of each synthetic function), the regions of high performance are localized in the same areas of the heatmaps. It is worth noticing that in Branin-2D, only $\rho$ values below 0.1 provide the best MFBO performance, similar to what was found in the synthetic functions experiment of a similar study\cite{judge2024}, and in previous studies on materials discovery\cite{Jacobs2023}. These results provide a qualitative understanding of the effect of cost and informativeness on the performance of MFBO. The general high-performance region is located around cheap and highly informative LF sources, and a good experimental setup for MFBO application should aim to find suitable LF sources.

\subsection*{Application of MFBO to chemistry and materials problems}
\begin{figure}[h]
\includegraphics[width=0.99\columnwidth]{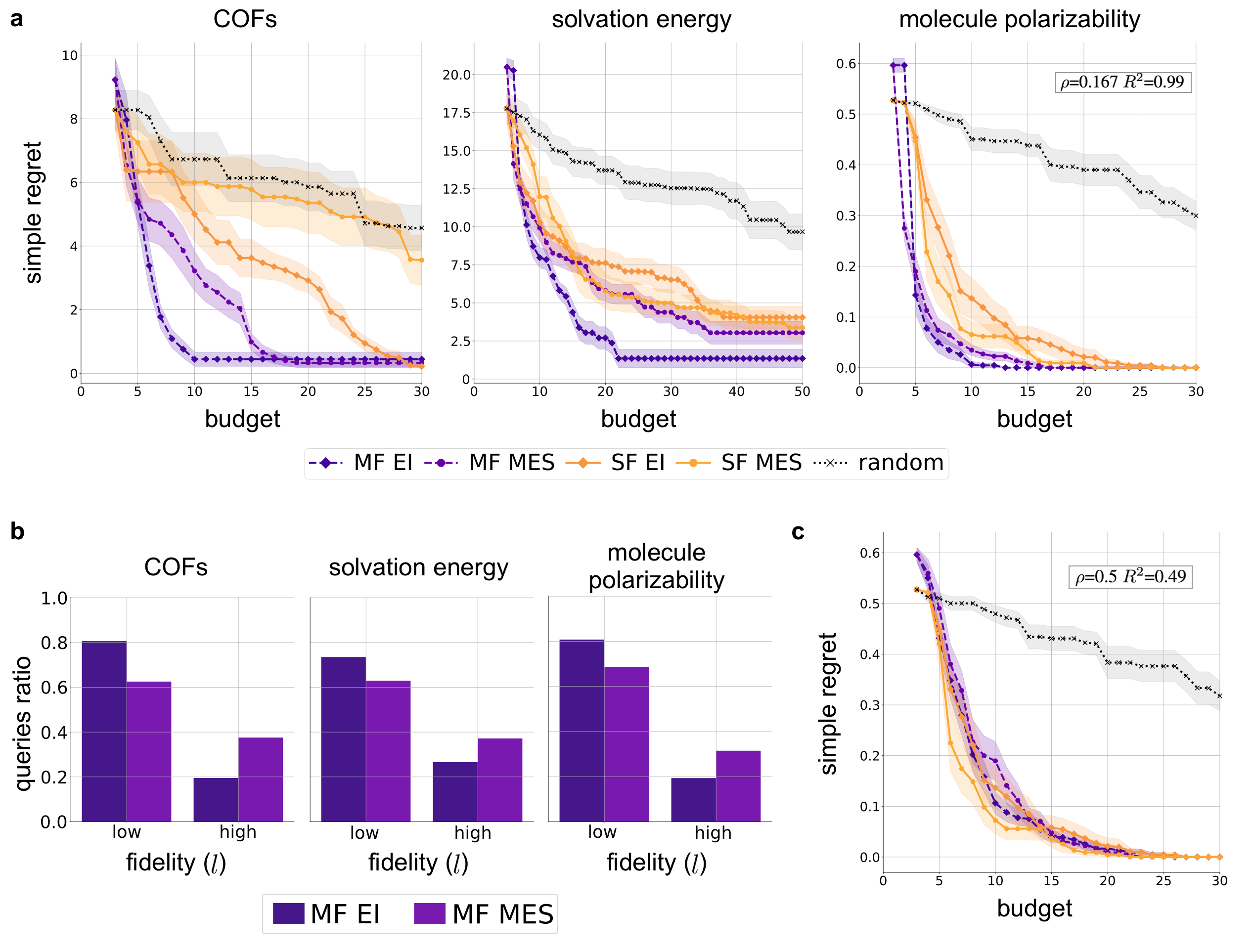}
\caption{Molecules and materials benchmark results. a) Optimization performance for the MFBO and the SFBO methods in each benchmark (see \hyperref[benchmarks]{Methods} for a description of each task). The maximum $\Delta$ in each case is 0.68, 0.59 and 0.56 for the COFs, solvation energy and polarizability  tasks, respectively. In all the cases, MFBO is able to achieve lower regrets than SFBO by spending a lower budget. b) Proportion of queries to each fidelity source in the MFBO setting. The MFBO method balances information gain and budget spent by querying more often the LF values (between 0.6 and 0.8 in the ratio of queries) while also using information from HF values. c) Experimental negative example. Molecule polarizability benchmark conditions were changed so the LF source was expensive and less informative. The resulting $\Delta$ is 0.07 and -1 for the MES and EI families respectively, showing how unfavorable conditions can override the MFBO advantage over SFBO in a real experimental setting. The reference budget unit cost corresponds to a single HF evaluation in all cases.}
\label{fig5}
\end{figure}

We evaluate three benchmarks in chemistry and materials science to study the performance of MFBO in more realistic experimental scenarios . The tasks correspond to scenarios where a cheap and informative LF source is available (see \hyperref[benchmarks]{Methods} for a detailed explanation of each benchmark). Figure \ref{fig5}a illustrates the results of MFBO and SFBO in these scenarios. In all cases, the MFBO method can get to lower regrets at a lower cost than the SFBO by leveraging the cheaper source of information at hand. In the COFs benchmark (problem settings: $\rho$ = 0.065, $R^2$ = 0.98), the MFBO gets to regrets close to 0 in an early phase of the optimization, whereas the SFBO only does it at the end of the optimization (the maximum $\Delta$ is 0.68). In the solvation energy benchmark (problem settings: $\rho$ = 0.1, $R^2$ = 0.88), MFBO also gets to regrets close to 0 at a lower budget than SFBO. This example also integrates computed values with experimental measurements to reduce the spent budget. In the polarizability benchmark (problem settings: $\rho$ = 0.167, $R^2$ = 0.99), the MFBO provides a maximum discount of 0.56 over the SFBO. In this case, the use of a cheap Hartree-Fock computation is successfully combined with the experimental measurements to find the molecule with the highest polarizability. Figure \ref{fig5}b shows the proportion of queries to each fidelity level in the different tasks. By distributing the available budget between the HF and LF levels, the MFBO methods can integrate the cheaper information of the LF approximations to improve the optimization results. The number of calls to each query depends on the specific acquisition function and task. However, in all cases the proportion of calls to the HF level is lower than 0.4, suggesting that this proportion can provide a threshold for a successful MFBO optimization. A previous study suggested a proportion of LF-HF points of 5:1 as the optimal case for good MFBO performance\cite{Jacobs2023}, which is similar to the 4:1 ratio observed in the COFs and polarizability benchmarks. In these real scenarios, the MFBO methods can exploit the low accuracy but cheap LF levels by querying them more often than the HF level.

We also include a negative example where the conditions of one of these benchmarks are changed to an unfavorable scenario and the associated original MFBO performance is overridden. We artificially change the LF conditions of the molecule polarizability benchmark ($\rho$ = 0.5 and $R^2$ = 0.49) to illustrate how $\Delta$ is reduced with respect to the previous case. Figure \ref{fig6}c shows the result for the optimization, where $\Delta$ is reduced from a maximum value of 0.56 to -1 in the EI case (that is, the MFBO gives worse result than the SFBO), providing no advantage over SFBO if compared with the previous favorable scenario (similar to the results described for the synthetic functions, see \hyperref[sec:failures]{Results}). This highlights the importance of using adequate LF sources, and how their incorporation can be detrimental otherwise\cite{mikkola23a}. The previous benchmarks show how the application of MFBO in suitable scenarios provides an advantage over standard SFBO cases and effectively decreases the optimization cost. By locating experiments where the cost and informativeness of the LF source are adequate, we can successfully translate the advantages of MFBO into a realistic experimental task.

% relate to previous studies (palizhati and jacobs)

\subsection*{Guidelines for MFBO in chemistry and materials}

\begin{figure}[h]
\includegraphics[width=0.99\columnwidth]{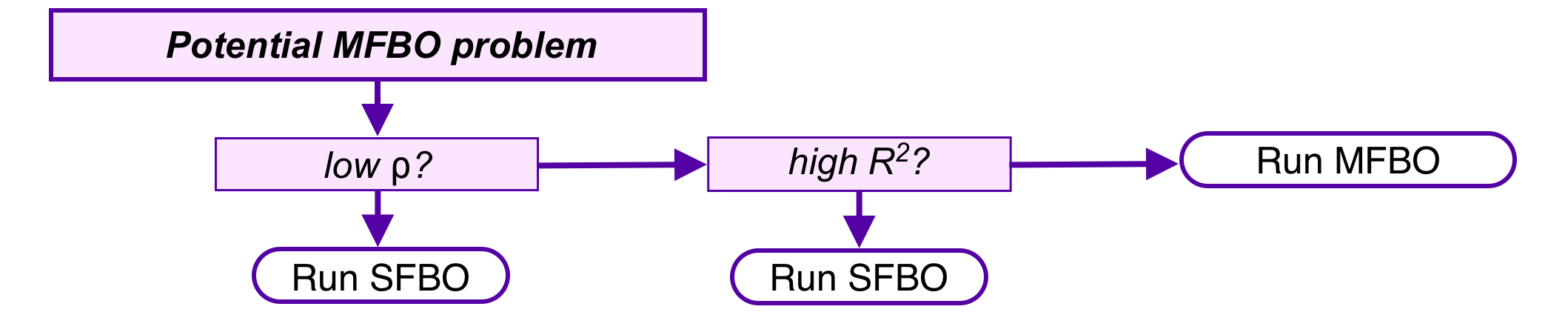}
\caption{Guidelines for MFBO application in chemistry and materials. Given a potential optimization problem where a cheaper source of information is available, the user should estimate if this source is cheap enough (e.g. $\rho$<0.1, based on this work) and informative enough (e.g. $R^2$>0.8, based on this work) before running MFBO . These parameters could be estimated on the initial data sampling or using previous experiments. If these conditions are not met, they should use SFBO otherwise.}
\label{fig6}
\end{figure}

We discuss and provide some recommendations to run MFBO in an experimental binary setting from the results obtained in this study. Our results highlight how the focus should be directed to the cost and informativeness of the LF source. Figure \ref{fig6} displays a guiding flowchart following the previous considerations. Using these conditions does not guarantee superior performance over SFBO, but provides a guiding principle for the successful application of MFBO in real scenarios.

In general, a user should first verify that the cost of the auxiliary experiment is cheap enough to run MFBO. All experimental benchmarks considered in this study employed a cost ratio below 0.167, with the synthetic benchmarks achieving optimal performance at $\rho = 0.05$. Notably, this observation aligns with prior findings in bandgap optimization, where the most effective MFBO performance was reported at a cost ratio of 5\% \cite{Jacobs2023}. This suggests that even though more expensive LF sources may also provide advantages, $\rho$ values around 0.05 can be a good starting point to try MFBO. In the same way, the user should check and estimate the degree of informativeness of the auxiliary experiment to decide if it provides a reasonable approximation of the real problem. The synthetic benchmark showed how uncorrelated ($R^2\leq 0$) LF functions always gave negative MFBO performance. In the experimental benchmark, all the LF $R^2$ values were above 0.85, and when artificially decreasing it to 0.49 and increasing $\rho$ from 0.167 to 0.5 in the polarizability case, the MFBO performance degraded. Previous work used LF values where $R^2$ = 0.84\cite{Jacobs2023}. Therefore we recommend $R^2$ values over 0.80 as a guiding principle to use informative LF sources.

%mention nonlinear correlation and the things that R3 mentioned
It is important to notice that informativeness measurement is difficult to estimate beforehand. In this case, we propose $R^2$ as a guiding metric, but other informativeness measurements could be employed (for instance, metrics that can capture nonlinear trends). Other studies have used the Pearson correlation coefficient\cite{judge2024} or the Matthew correlation coefficient \cite{macdonald2025_mfbo} to estimate the informativeness of the LF sources. Also, the cost is arbitrary depending on the user and experiment, as different setups may prioritize some costs over others. For example, experiment time is the most intuitive cost measurement, but also experiment price or any other relevant quantity could be used.
In addition, it may be difficult to accurately estimate these parameters when starting the optimization from scratch. A potential option in a realistic scenario is to use the initial sampled points to evaluate the cost and informativeness of the LF source with respect to the sampled MF by running a simple regression task to check if the MF surrogate improves the SF predictions. This alternative estimation is left for future work.

Despite all these challenges, the idea holds: the user should reflect if the conditions are met for MFBO to avoid unreliable LF sources\cite{mikkola23a} or long-term performance degradation of the MFBO method\cite{dovonon2023longrunbehaviourmfbo} in order to successfully apply MFBO in real scenarios.

\section*{Discussion}

Multi-fidelity Bayesian optimization (MFBO) offers a promising approach to reducing costs in optimization by leveraging inexpensive, approximate sources of information. In the chemical sciences, MFBO can accelerate optimization while maintaining cost efficiency. However, its utility is not universal; in some cases, it may prove less effective or even counterproductive\cite{mikkola23a}. To address these concerns, we conducted a comprehensive study to determine under which conditions MFBO outperforms standard single-fidelity Bayesian optimization (SFBO).

Building on previous work in the field, we introduce two key metrics for comparison: a standardized regret, $r$, and a discount metric, $\Delta$ (see \hyperref[sec:metrics]{Metrics}). Our analysis of synthetic benchmarks demonstrates that MFBO performance is highly dependent on the characteristics of the low-fidelity (LF) source. Specifically, not all LF conditions are conducive to the effective use of MFBO. To systematically identify favorable conditions, we performed a grid-based exploration of LF cost and informativeness against the discount metric, $\Delta$. The results indicate that both $\rho$ and $R^2$ significantly influence the discount, with the most favorable outcomes achieved when the LF source is both highly informative and low-cost. The method was validated using three real-world experimental benchmarks in molecular and materials discovery, demonstrating that MFBO consistently outperforms SFBO in reducing optimization costs, provided the LF source is sufficiently inexpensive and informative. Based on these findings, we developed a flowchart to guide the application of MFBO (see Figure \ref{fig5}).

Our results resonate with previous studies suggesting a favorable cost ratio of 0.05\cite{Jacobs2023} and showing how informative sources are required for MFBO to outperform SFBO\cite{judge2024}. In addition, the optimal LF/HF query proportion of 4:1 observed in the experimental benchmarks also aligns with previous observations\cite{Jacobs2023} that found 5:1 as the best ratio, although we use an adaptive acquisition function instead of a greedy heuristic. Although the modelling was not the main focus of the study, other works used different surrogates like one-hot encoded GP\cite{Palizhati2022} or a random forest\cite{Jacobs2023}, and comparing the MFBO and SFBO performance of all the surrogates could be a future research direction to complement the guidelines. In addition, model uncertainty calibration may also play a key role during the acquisition stage, and studying its effect on MFBO performance would help to clarify how its effect interacts with poor or high LF informativeness.

Our work provides a structured decision-making framework for determining the applicability of MFBO. While our analysis focused on a specific surrogate model and two acquisition functions, the approach is extensible to other surrogate models and acquisition function families. Future research will expand this framework to include models such as Bayesian neural networks (BNNs)\cite{li2024studybayesianneuralnetwork}, and explore additional experimental applications, evaluating the impact of varied feature spaces or additional fidelity levels. This study offers valuable insights for practitioners seeking to integrate MFBO into their experimental workflows, paving the way for more routine application of MFBO in chemical and materials optimization.
% would it be good to reflect (as the NeurIPS reviewer suggested) if the scenarios that we are using as a benchmark are realistic? (e.g.: high R2 is observed in many cases)

\section*{Methods}

\subsection*{Multi-fidelity Bayesian Optimization}\label{MFBO explained}
BO uses a probabilistic surrogate model and a selection policy (acquisition function) to optimize black-box problems. These problems are normally expensive to query, and BO accelerates the finding of optimal points by minimizing the number of calls to the target function. The optimization problem can be formulated as 

\begin{equation*}
\argmax_{{x} \in \mathcal{X}} f({x})
\end{equation*}

where \textit{f} is the target black-box function and $\mathcal{X}$ is the space of all possible input candidates. Considering $\mathcal{D} = [(x_1, y_1), (x_2, y_2),...(x_n,y_n)]$ a dataset containing pairs of inputs-outputs $({x}_n,y_n) \in \mathbb{R}^d \times \mathbb{R}$ of our problem, we can use a surrogate model to learn the relationships between the data in the form $ y_i = f({x}_i) + \epsilon$, with $\epsilon \sim \mathcal{N}(0,\sigma_{\epsilon}^2)$. The surrogate is commonly a Gaussian Process (GP), a non-parametric model that provides uncertainty quantification \cite{Rasmussen2006}. GPs are usually defined as $f(x) \sim \mathcal{GP} (0, k({x}, {x'}))$, placing a zero-mean prior and a covariance given by a kernel function $cov[f({x}), f({x'})] = k({x}, {x'})$ that measures the similarity between inputs. Given a dataset $\mathcal{D}$, the predictive mean and variance are given by the following expressions

\begin{equation*}
\begin{split}
    \mu({x}) &= k_{x}(K + \sigma_{\epsilon}^{2}I)^{-1}{y} \\
    \sigma^{2}({x}) &= k({x}, {x}) - {k}_{x}(K + \sigma_{\epsilon}^{2}I)^{-1}{k}_{x}^\top
\end{split}
\end{equation*}

where ${k_x} = [k({x}, {x_1}), ..., k({x}, {x_n})]^\top \in \mathbb{R}^n$, $K \in \mathbb{R}^{n\times n} = (k({x_i}, {x_j}))_{1\leq i, j\leq n}$, ${y} = [y_1,..., y_n]$, \textit{I} is the $n \times n$ identity matrix, and $\sigma^2_{\epsilon}$ is the variance of the observation noise. In order to optimize the target problem, after training the surrogate with the available data, a decision is made to select the next points by maximizing the acquisition function. Acquisition functions are heuristics that decide the next point to query by balancing exploration and exploitation. Exploration promotes sampling in regions of high uncertainty to improve the model's understanding of the search space, while exploitation prioritizes regions near the current best observations to refine promising solutions. Each acquisition function family offers different tradeoffs between these two terms. The next inputs are given by

\begin{equation*}
    x_{n+1} = \argmax_{{x} \in \mathcal{X}} \texttt{acqf}({x; \mathcal{D}})
\end{equation*}
where $\texttt{acqf}$ is the acquisition function, which uses the surrogate trained on the sampled dataset $\mathcal{D}$ to select the next query among the candidates. Common acquisition functions include Expected Improvement (EI)\cite{Jones1998-ei}, Maximum Entropy Search (MES)\cite{maxvalueentropysearchefficient} and Knowledge Gradient (KG)\cite{frazier2018tutorialbayesianoptimization}.

Our multi-fidelity BO extends the previous setup by considering an extra input parameter for the surrogate model, the fidelity \textit{l}. This parameter indicates which source of information the other inputs correspond to (e.g., high fidelity, HF, or low fidelity, LF). In our case, we use a discrete fidelity setting, where $l \in [l_0, l_1,...l_m]$ and $l_m$ corresponds to the highest fidelity among the $m$ levels. In all cases, we limit the analysis to two levels of fidelity ($l_0$ and $l_1$ and $m=1$)
, but the methodology can be extended to more levels (see Supplementary Information \ref{catalysis_task} for an example of a 3-fidelities regression task). Additionally, these models assume that the cost only depends on the fidelity level $l$ and not on the search space $x$. We employed a modified GP proposed in \cite{wu2020} that extends the previously described surrogate modelling to the multi-fidelity space of $\mathbb{R}^{n+1}$ dimensions, chosen due to its popularity amongst MFBO applications\cite{wu2020, Gantzler2023-fj, mikkola23a}. The modified kernel is defined as $k(({x},l),  ({x'}, l'))$, where fidelity is introduced by defining a separate kernel to model the input space, and a kernel to model the correlation and interaction between the different fidelity levels. The mathematical expression is:

\begin{equation*}
    \begin{split}
        k\left(\left({x},l\right),  \left({x'}, l'\right)\right) &= k_{\rm I}({x}, {x'}) \times k_{\rm IS}(l, l')\\
        k_{\rm I}({x}, {x'}) &= \exp\left( -\frac{1}{2} \sum_{i=1}^{d} \lambda_i^{-1}(x_i - x_i')^2 \right) \\ 
        k_{\rm IS}(l, l') &= c + (1-l)^{1+\delta}(1-l')^{1+\delta}\\
    \end{split}
\end{equation*}

Given a dataset of inputs with their associated fidelities and outputs, the model can be trained using standard maximum marginal log-likelihood optimization to obtain the optimal hyperparameters $\lambda_i$, $c$ and $\delta$. In the experiments, we set the $l$ values of the HF and LF levels to $\frac{2}{3}$ and $\frac{1}{3}$ respectively following previous studies that used this kernel in a binary setting\cite{Gantzler2023-fj}. The choice of $l_{0}$, and $l_{1}$ values was done to ensure the c and the exponents of the $k_{IS}$ have an impact on the kernel value, which may not be the case if other pair values of l and l' were chosen such as 0/1.

In the multi-fidelity case, acquisition functions must compute the information gained over the cost of a query at a given fidelity level. The previous expression for the acquisition function maximization can be therefore generalized as $\texttt{acqf}(x, l) = \texttt{acqf}(x)\cdot \texttt{cost}(l)^{-1}$. The next pair of input and fidelity query is given by 
\begin{equation*}
    x_{n+1}, l_{n+1} = \argmax_{(x,l) \in \mathcal{X} \times [m]} \texttt{acqf}(x, l; \mathcal{D})
\end{equation*}

In this setting, the acquisition function selects the input-fidelity pair by also considering the cost of each fidelity. In the experiments, we use two families of acquisition functions, with their single and multi-fidelity versions: Expected Improvement (EI)\cite{Huang2006-jm} and Maximum Entropy Search (MES)\cite{takeno20a}. 

The expression of the multi-fidelity EI function is

\[
\begin{adjustbox}{max width=\textwidth}
$
    \begin{aligned}
        x_{n+1}, l_{n+1} &= \argmax_{(x,l) \in \mathcal{X} \times [m]} \mathbb{E} \left[ \max \left[ 0, Y^{m} (\mathbf{x}) \Big| \mathcal{D}_{n} - \hat{y}^{m*}_{n} \right] \right] 
        \times \operatorname{corr} \left[ Y^{(\ell)} (\mathbf{x}) \Big| \mathcal{D}_{n}, Y^{(m)} (\mathbf{x}) \Big| \mathcal{D}_{n} \right]
        \times \frac{1}{\rho}
    \end{aligned}
$
\end{adjustbox}
\]

Where $\mathcal{D}_n$ is the dataset acquired at a step \textit{n} of the optimization. The first term is the EI in the objective property of the HF simulation, and balances exploration and exploitation by favoring candidates with high and/or uncertain predicted objective. This term is computed as the expectation of the predicted HF values $Y^{m} (\mathbf{x}) \Big| \mathcal{D}_{[n]}$ greater than the best HF value observed so far, $\hat{y}^{m*}_{[n]}$. The second term is the correlation between the target property of \text{x} under the \textit{l} fidelity and the high-fidelity $l_{m}$, which penalizes uncorrelated low-fidelity values. The last term is the inverse of the cost ratio between the high fidelity source $l_{m}$ and the fidelity \textit{l}, which rewards cheaper samples. The composed function rewards candidates that have a high utility per cost for the target objective. 

The expression of the multi-fidelity MES function is

\begin{equation*}
        x_{n+1}, l_{n+1} = \argmax_{(x,l) \in \mathcal{X} \times [m]} \left(\mathbb{H}(y_m \mid x,\mathcal{D}_{n}) - \mathbb{E}_{f^{*}\mid x,\mathbb{D}}[\mathbb{H}(y_m \mid f^{*}, x,\mathcal{D}_{n})]\right) \times \frac{1}{\rho}
\end{equation*}

Where $\mathbb{H}$ is the entropy, which measures uncertainty or the information content of a random variable or distribution. The first term $\mathbb{H}(y_m |x,\mathcal{D}_{n})$ represents the entropy of the function value at the highest fidelity level $l_{m}$. It reflects the uncertainty about the objective function at the candidate point \textit{x} before making a new 
observation. The second term $\mathbb{E}_{f^{*}\mid x,\mathbb{D}}[\mathbb{H}(y_m\mid f^{*}, x,\mathcal{D}_{n})]$ denotes the expected entropy of the posterior distribution of the highest fidelity function value at point \textit{x}, conditioned on a hypothesized realization of the global optimum $f^{*}$. This conditional entropy indicates the uncertainty after observing the hypothetical optimal function value. The difference between these two entropy terms is known as the information gain, which measures how much observing a particular point at fidelity \textit{l} is expected to reduce uncertainty about the optimal solution.
The third term is the inverse of the cost ratio between the high fidelity source and the fidelity \textit{l}. MES rewards candidates that bring a reduction of high-fidelity entropy scaled by their respective cost. 

We also initially tested Knowledge Gradient (KG)\cite{wu2020}, but we discarded it due to its computational expense.

In all cases, the experiments are run for 20 independent seeds. We assign 10\% of the total (pre-selected) optimization cost to the initial sampling, which we refer to as sampling cost (the unit of cost in this work is a HF evaluation). In all cases, the selected budget value is selected to achieve low regrets in the SFBO case. Then, we assign 50\% of this budget to HF samples and 50\% to low fidelity samples (the number of HF and LF samples depend on the relative cost of each one). Given that we define cost in terms of HF evaluation units,  if a sampling cost of 4 is used, we take 2 HF samples and 2/$\rho$ LF samples. For continuous spaces, we employ a Latin Hypercube Sampling (LHS) strategy. In the case of categorical benchmarks, we adopt a methodology similar to the one described in previous work\cite{Gantzler2023-fj}. This approach involves the random selection of an initial example, followed by the sequential selection of the most spatially distant samples relative to those already chosen (Furthest Point sampling).

In the synthetic functions, the total budget corresponds to 50 HF queries. In the chemistry and materials benchmarks, the total budget corresponds to 30 HF queries for the COFs and polarizability, and 50 HF for the solvation energy task. Data collection is sequential. All the experiments are run using BOTorch \cite{balandat2020botorch} as a standard Bayesian Optimization framework. 

\subsection*{Metrics}\label{sec:metrics}
Recent work in MFBO for materials discovery has seen the emergence of several metrics to evaluate method performance. To measure the progress of the optimization campaign, regret\cite{Fare2022-yi, Gantzler2023-fj}, Active Learning Metrics (ALM)\cite{Palizhati2022} and campaign efficiency\cite{Jacobs2023} have been proposed. In the ML literature, regret is the most common metric, although its specific implementation for MFBO cases is not defined\cite{nips2016_MFUCB, wu2020, mikkola23a, Folch2023-nn}. To measure absolute MFBO performance, that is, how a specific MFBO run compares to its SFBO counterpart under specific experimental conditions, Acceleration Factor (AF) and Enhancement Factor\cite{Palizhati2022}, cost difference to discover materials in the 99th best percentile\cite{Fare2022-yi} and MF advantage\cite{Jacobs2023} have been used. Importantly, MFBO performance is also dependent on the available budget, and this absolute metric has to somehow capture this dependency. Only some works have explicitly mentioned \cite{Jacobs2023, Gantzler2023-fj} or studied\cite{Fare2022-yi} how low-fidelity informativeness affects the final result, quantifying it using the $R^2$ or correlation coefficient between the HF and LF source, respectively. Last, some recent works highlighted the need of better metrics to assess MFBO performance \cite{dovonon2023longrunbehaviourmfbo, judge2024}. To solve this lack of standardization, and building on the previous works, we propose two key metrics - MFBO regret (r) and discount ($\Delta$) to study MFBO performance under different scenarios. We also define $\rho$ and $R^2$ as the metrics to characterize the LF source.

\subsubsection*{High-fidelity regret calculation in MFBO setting}

\begin{figure}
    \centering
    \vspace{-1em}
    \includegraphics[width=0.65\columnwidth]{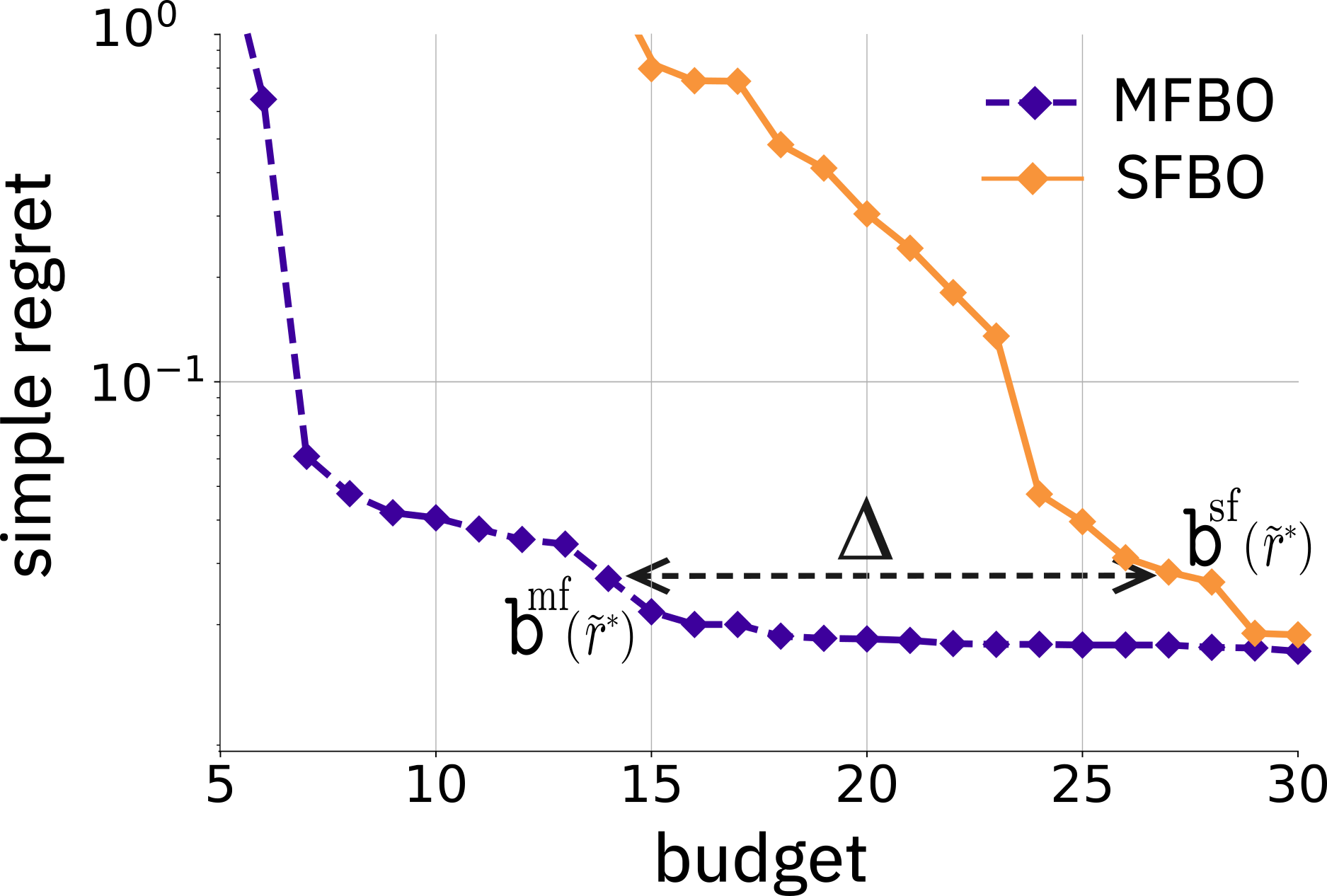}
    \caption{Standardized metrics to compare MFBO and SFBO. Simple regret traces are computed using algorithm \ref{regret algo} and provide a direct comparison between the optimization progress in each method by computing the equivalent number of steps from SFBO in the MFBO trace. Discount ($\Delta$) reflects the savings in optimization cost by computing the normalized difference in budget spent by each algorithm ($\texttt{b}^{\rm sf}(\tilde{r}^*)$ and $\texttt{b}^{\rm mf}(\tilde{r}^*)$) to find $\tilde{r}^*$ (horizontal arrow, see equation \ref{eqn:delta}). The plot corresponds to the average trace of the MFEI and SFEI methods on an example favorable Branin-2D case (see Figure 3), and it is zoomed in to focus on the $\Delta$ score.}
    \vspace{-0.5em}
    \label{fig2}
\end{figure}

We define a standardized simple regret and a discount metric to quantify the performance of MFBO over SFBO along the optimization. This metric aims to inform how the MFBO optimization progresses with respect to a known reference single-fidelity source. This approach is similar to the idea of ALM introduced in \cite{Palizhati2022} (note that it is an "after-the-fact" metric, meaning that the best candidate must be known beforehand, although this is common to all the benchmarks). Let $f$ be the HF target problem. At a given step $t$ of the SFBO run, the simple regret is defined as $r_{t} = f^* - f^*_t$, where $f^*$ is the known global optimum and $f^*_t$ is the best value of $f$ found so far. The simple regret is computed only at the highest fidelity level in the multi-fidelity runs because this level is the real optimization target and the true reference to measure the method's performance. To compare multi-fidelity and single-fidelity runs it is necessary to find the corresponding high-fidelity simple regret in the multi-fidelity setting, which can query both the low-fidelity or high-fidelity at a given step. We define a standard algorithm to identify the corresponding HF simple regret at a given budget of an MFBO campaign. The algorithm aligns the ${r}$ values from the MFBO to a common cost scale given by the SFBO cost steps. This approach makes it possible to compare the performance of the MFBO method at each step of the associated SFBO run. 
Algorithm \ref{regret algo} explains in detail the procedure to compute simple regret both for single-fidelity and multi-fidelity runs.
Even if in our work we only evaluate two sources of fidelity, the advantage of this regret computation is that it allows to incorporate as many discrete LF sources as the user wants. A potential disadvantage may be the need of a reference optimization case (the SFBO run in this case), but we consider this acceptable as the final objective of the study is to decide when MFBO is better than SFBO.

In the plots, we compute the average $r$ and standard deviation of an optimization run using the proposed algorithm from all the different repeats. Figure \ref{fig2} shows a graphical explanation of the associated regret traces depicted in orange and purple which have an equal number of datapoints. 

\subsubsection*{Discount as a utility metric for MFBO}
We also propose a discount $\Delta$ metric to estimate the savings provided by the MFBO method compared to the SFBO (see figure \ref{fig2} for a graphical explanation of this metric). This metric is inspired by the Acceleration Factor (AF) and Active Learning Metrics (ALM) previously proposed and used in MFBO work\cite{Palizhati2022}.
$\Delta$ reflects the difference in budgets required to reach a reference regret value between the multi-fidelity and single-fidelity methods. This metric can be viewed as a combination of the simple regret and the total spent budget, reflecting how MFBO results are an interplay of the HF value optimization and the total budget spent. Specifically, it is computed over two individual SF and MF runs as follows:

\begin{equation}
    \Delta (\tilde{r}^*) = \dfrac{\texttt{b}^{\rm sf}(\tilde{r}^*) - \texttt{b}^{\rm mf}(\tilde{r}^*)}{\texttt{b}^{\rm sf}(\tilde{r}^*)}, \text{where } \tilde{r}^* = r^{\rm sf}_{\rm max} - (r^{\rm sf}_{\rm max} - r^{\rm sf}_{\rm min}) \tau.
\label{eqn:delta}
\end{equation}

% tilde r is the target sf regret. This is the regret from the single fidelity run that you are targetting (e.g.: 0.9 of the final regret) Tau accounts for this difference, and allows to measure the progress of the optimization at different times (notic that the 
$\tilde{r}^*$ is the \textit{corrected best single-fidelity regret}, and $r^{\rm sf}_{\rm max}$ and $r^{\rm sf}_{\rm min}$ are the maximum and minimum regrets in the SFBO run, and $\tau$ is a factor that quantifies the amount of total reduced regret that the user wants to sacrifice in order to get a good MFBO performance. The $\tau$ correction accounts
for situations where the MFBO method gets to low regrets faster than the SFBO at the beginning
of the optimization, but its performance plateaus in the long term as noted in\cite{dovonon2023longrunbehaviourmfbo}. This correction also reflects the dependence of MFBO performance on the available budget, as mentioned in previous work\cite{Jacobs2023}.$\tau$ refers to the slack a user will give with respect to the converging regret of SFBO, or how good the user wants their output to be with respect to the minimum regret achievable by SFBO.

We report discounts with $\tau$ = 0.9, although we also investigate the discount obtained for several values of $\tau$ in the synthetic functions benchmark (see SI section \ref{si: discount3d}. A positive $\Delta$ indicates cost savings with MFBO, meaning it achieves the minimum regret with a lower budget than SFBO. Conversely, a negative $\Delta$ implies that MFBO was more expensive than SFBO. If the MFBO method is unable to reach $\tilde{r}^*$, $\Delta$ is set to -1. Note that equation \ref{eqn:delta} compares individual runs started at the same seed as displayed in Figure \ref{fig2}, and we report the average $\Delta$ over all the seeds onwards.

\subsection*{LF level cost and informativeness estimation}\label{sec:costandinfo}
We use cost ratio and informativeness metrics to compute the characteristics of the low-fidelity approximation with respect to the high-fidelity. Cost is characterized by the cost ratio $\rho$, which is obtained by dividing the LF cost and the HF cost. Informativeness is characterized by the $R^2$ of the LF approximation. To compute this metric, we uniformly sample 100 points for the given problem and extract their associated values at the HF and LF levels. Then, the $R^2$ between a linear fitting of the LF prediction and the true HF value is computed. Supplementary Information \ref{R2problems} shows the plots with the sampled points and the computed $R^2$ of each of the problems used in this study.

\subsection*{Synthetic functions}\label{synthetic_funcs}
Synthetic functions are simulated black-box problems where a mathematical expression is used to generate points from a surface to be optimized. In the multi-fidelity case, there are lower accuracy approximations of the target function at a lower cost. These can either have a fixed expression or be generated by biasing the original function using a given parameter $\alpha$. In our case, we use the Branin\cite{mikkola23a} function previously used in MFBO works. We also modify a Park function used in MFBO\cite{Xiong2013-park} by incorporating a parameter $\alpha$ to modulate the bias with a similar behaviour to the Branin case. The expressions of the functions can be found in the Supplementary Information \ref{sfs_expressions:A2}, and a visualization of the $\alpha$ parameter's effect on the Branin function (figure \ref{fig:branin_alpha}). The cost of querying the high-fidelity level is set to a value of 1, and the cost of the low-fidelity approximations is set to a fraction of this value.

\subsection*{Chemistry and materials design benchmarks}\label{benchmarks}
We use or adapt previously reported benchmarks for real experimental optimization problems in the chemistry domain. The Covalent Organic Frameworks (COFs) dataset was used in a previous work on MFBO\cite{Gantzler2023-fj}, and consists of 608 candidate COFs encoded in a 14-dimensional vector accounting for their composition and crystal structure. The high-fidelity simulation uses a grand canonical Markov chain Monte Carlo simulation to compute the adsorption of Kr and Xe in the material, and the associated Xe/Kr selectivity. This simulation is expensive, with an average running time of 230 minutes. In this case, a low-fidelity approximation of the selectivity can be obtained using Henry's law with a classical force field, reducing the computing time to 15 minutes, giving a $\rho$ of 0.065 for this problem. 

The second benchmark is extracted from the FreeSolv library\cite{Mobley2014-xu}.
It comprises 641 molecules, encoded using RDKit 2D-descriptors and reduced to a 10-D vector using PCA. The high-fidelity is the experimental free solvation energy, whereas the low-fidelity is the computed solvation energy using molecular dynamics (MD). $\rho$ is set up to 0.1 in this case, based on an estimation of the difference between running a solvation measurement and an MD simulation.

The last benchmark is derived from the Alexandria library \cite{Ghahremanpour2018-alexandria}, and it was used in previous work on MFBO for materials discovery\cite{Fare2022-yi}. It comprises 1134 molecules encoded using RDKit 2D-descriptors \cite{rdkit} and reduced to a 10-D vector using PCA. The high-fidelity is the experimental polarizability, whereas the low-fidelity is the computed polarizability at the Hartree-Fock 6-31G+ level of theory. $\rho$ is set up to 0.167 in this case following the previous study.

\section*{Data availability}
The COFs benchmark data is available at \url{https://github.com/SimonEnsemble/multi-fidelity-BO-of-COFs-for-Xe-Kr-seps.}, and it was extracted from \cite{Gantzler2023-fj}. The polarizability dataset is available at \url{https://zenodo.org/records/1004711}\cite{Ghahremanpour2018-alexandria}. It was used in previous MFBO work\cite{Fare2022-yi}. The solvation energy dataset is available at \url{https://escholarship.org/uc/item/6sd403pz} and is based on \cite{Mobley2014-xu}. 

\section*{Code availability}
The code is available in this Github repository \url{https://github.com/atinary-technologies/chem-MFBO.git} (ref. \cite{Sabanza_Gil2025-we}). It includes the instructions to run all the experiments described in this work.

\section*{Acknowledgements}

This work was created as part of NCCR Catalysis (grant number 180544), a National Centre of Competence in Research funded by the Swiss National Science Foundation. V.S.G acknowledges support from the European Union’s Horizon 2020 research and innovation program under the Marie Skłodowska-Curie grant agreement N° 945363.

\section*{Author contributions}
V.S.G: conceptualization, methodology, software, investigation, visualization, writing - original draft. R.B.: conceptualization, methodology, software, writing - original draft. D.P.G.: conceptualization, methodology, software, writing - original draft. J.S.L: supervision, funding acquisition, resources, writing-review \& editing. J.M.H-L: conceptualization, writing-review \& editing. P.S.: supervision, funding acquisition, project administration, resources, writing-review \& editing. L.R.: conceptualization, project administration, funding acquisition, supervision, resources, writing-review \& editing. 

\section*{Competing interests}
L.R. is the CTO of Atinary Technologies, a company providing software and solutions for self-driving labs. D.P.G and R.B. are full-time employees at Atinary Technologies.

\bibliography{sn-bibliography}% common bib file

\begin{appendices}
\newpage

\section{Supplementary information}\label{secA1}



%% Appendices may be used for helpful, Supplementary or essential material that would otherwise 

\subsection{Preliminary investigation on the multi-fidelity kernel definition}

We compare two different GP models that can be used in multi-fidelity tasks. Each model uses a specific kernel to model the covariance between fidelities.
These kernels are part of the SingleTaskMultiFidelityGP and the MultiTaskGP models in BoTorch\cite{balandat2020botorch} respectively, and they are used to model the fidelities of the samples. 

The Downsampling kernel models the fidelity interaction in the SingleTaskMultiFidelityGP, and it has been already described in the Metrics section. The Index kernel models the fidelity interaction in the MultiTaskGP, and it is defined as:

\[
k(l, l') = \left( BB^\top + \operatorname{diag}(\mathbf{\sigma}) \right)_{l,l'}
\]

if there are two fidelities and rank = 1 (current case), $B = \begin{bmatrix}\alpha \\ \beta
\end{bmatrix}$ and 
\[
k(l, l') =
\begin{bmatrix}
\alpha^2 + \sigma & \alpha \beta \\
\alpha \beta & \beta^2 + \sigma
\end{bmatrix}.
\]

% This can be represented as a lookup table:
% \[
% \begin{aligned}
% k(l, l) &= \alpha^2 + \sigma \\
% k(l, l') &= \alpha \beta \\
% k(l', l') &= \beta^2 + \sigma
% \end{aligned}
% \]
% where \(l\) and \(l'\) denote the indices for the two fidelities.

% The Downsampling kernel is defined as:
% \[
% k(l, l') = c + (1 - l)^{1 + \delta} (1 - l')^{1 + \delta}
% \]
% If there are only two fidelities, 
% \[
% k(l, l') =
% \begin{bmatrix}
% c + (1 - l)^{2(1 + \delta)} & c + (1 - l)^{(1 + \delta)}(1 - l')^{(1 + \delta)}\\
% c + (1 - l)^{(1 + \delta)}(1 - l')^{(1 + \delta)} & c + (1 - l')^{2(1 + \delta)}
% \end{bmatrix}
% \]
% This is equivalent to the lookup table 
% \[
% \begin{aligned}
% k(l, l) &= c + (1 - l)^{2(1 + \delta)} \\
% k(l, l') &= c + (1 - l)^{(1 + \delta)}(1 - l')^{(1 + \delta)}\\
% k(l', l') &= c + (1 - l')^{2(1 + \delta)}
% \end{aligned}
% \]

This results in an similar structure between the two fidelity kernels, where there are only three possible fidelity interactions, and the kernel scales the result of the input kernel depending on the corresponding interaction value. 

\begin{figure}
    \centering
    \includegraphics[width=0.99\linewidth]{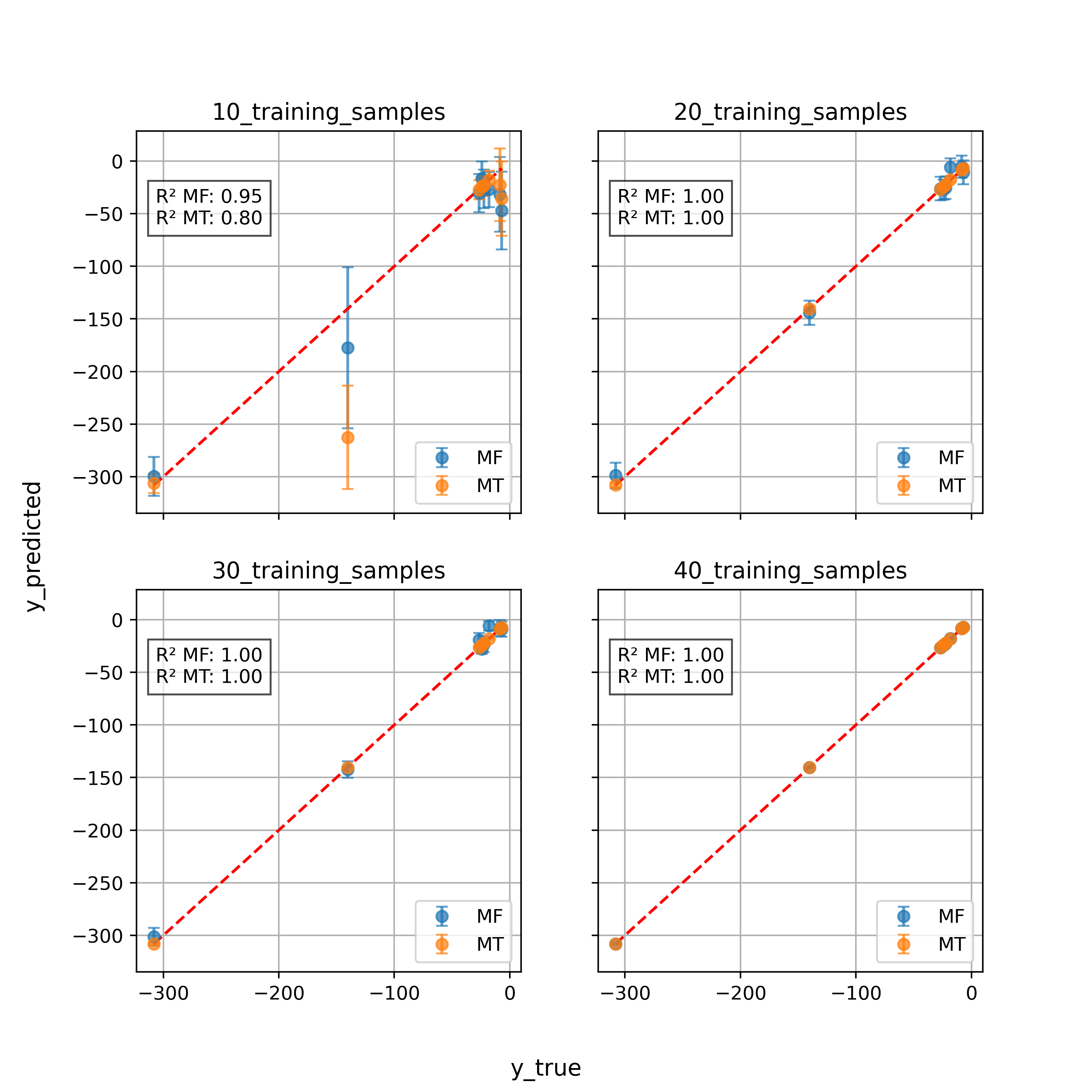}
    \caption{Regression performance of each surrogate with increasing training data for the Branin task. MF: default multi-fidelity model with Downsampling kernel, MT: MultiTask GP model with Index kernel.}.
    \label{fig:mf_mt_regression}
\end{figure}

\begin{figure}
    \centering
    \includegraphics[width=0.5\linewidth]{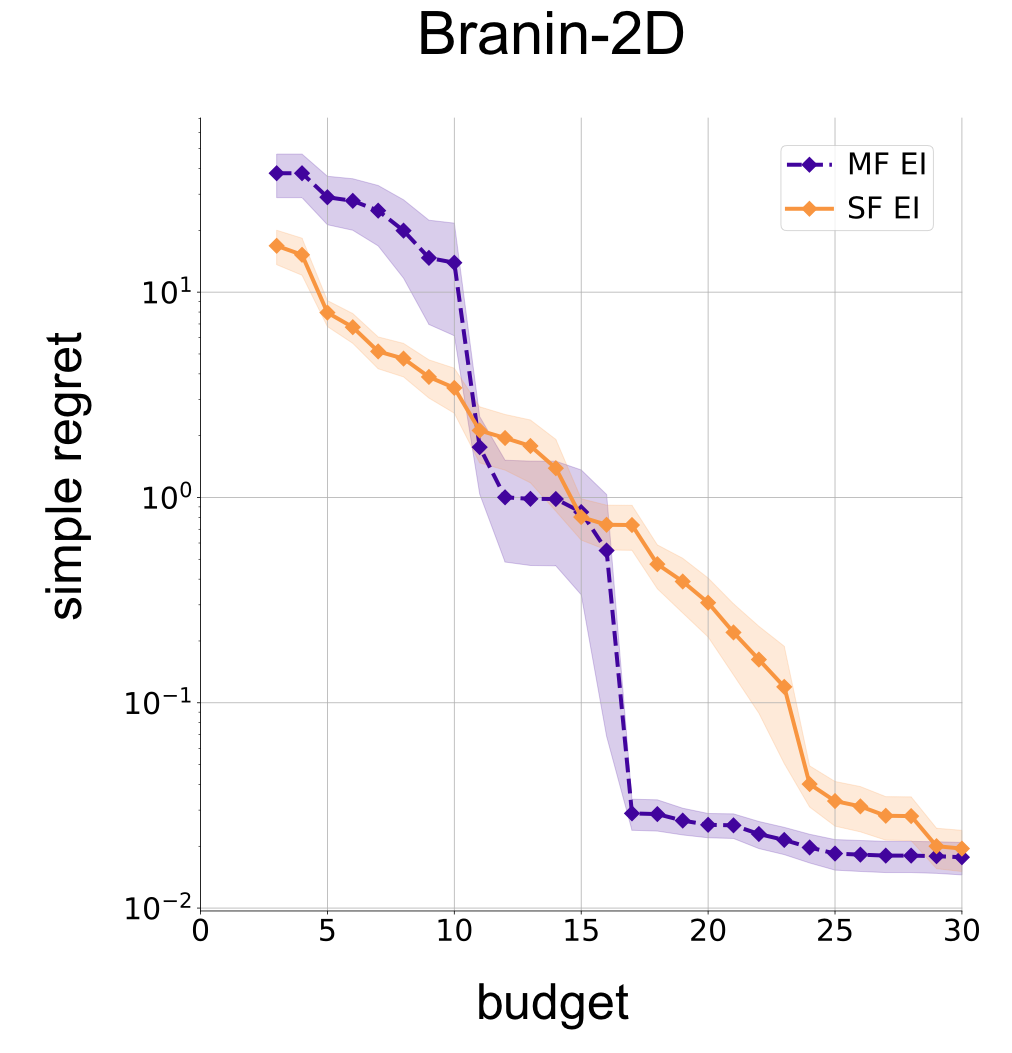}
    \caption{Results for BO loop using the MultiTask GP model with Index kernel and Expected Improvement on Branin-2D function under favourable conditions. The measured discount is $\Delta$=0.29}
    \label{fig:mt_branin}
\end{figure}

We tested the performance of each model to compare the effect of the kernel. In a regression task, we sample 8 high-fidelity points in the Branin-2D function as a test set and train each surrogate with increasing training points (simulating the progressive acquisition of samples in the active learning loop). Models were trained with 10, 20, 30 and 40 samples (with a 50-50 ratio of high and low fidelity points) and the model performance was calculated using the $R^{2}$ computed on the 8 test set points. Figure \ref{fig:mf_mt_regression} shows how both models have a similar performance in the high-data scenarios, with the default multi-fidelity kernel offering slightly better performance in the low-data scenario (10 training samples). For the MFBO case, the MT kernel was tested in the same settings as the favourable scenario of figure \ref{fig3}. Figure \ref{fig:mt_branin} shows how the performance of the model is similar to the default model, obtaining a slightly lower $\Delta$ (0.29).

\newpage
\subsection{Synthetic functions expressions}\label{sfs_expressions:A2}

\textbf{Branin-2D}
\begin{equation}
f(\mathbf{x},\alpha) = \left(
x_{2} - \left( \frac{5.1}{4\pi^{2}} - 0.1\left(1 - \alpha\right) x^{2}_1 + \frac{5}{\pi}x_1 - 6\right)^{2} + 10 \left(1 - \frac{1}{8\pi}\right) \cos{x_1}
\right) + 10 
\end{equation}
defined over $[-5, 10] \times [0, 15]$, and $\alpha \in [0, 1]$ is the bias term. \newline

\begin{figure}
    \centering
    \includegraphics[width=0.95\linewidth]{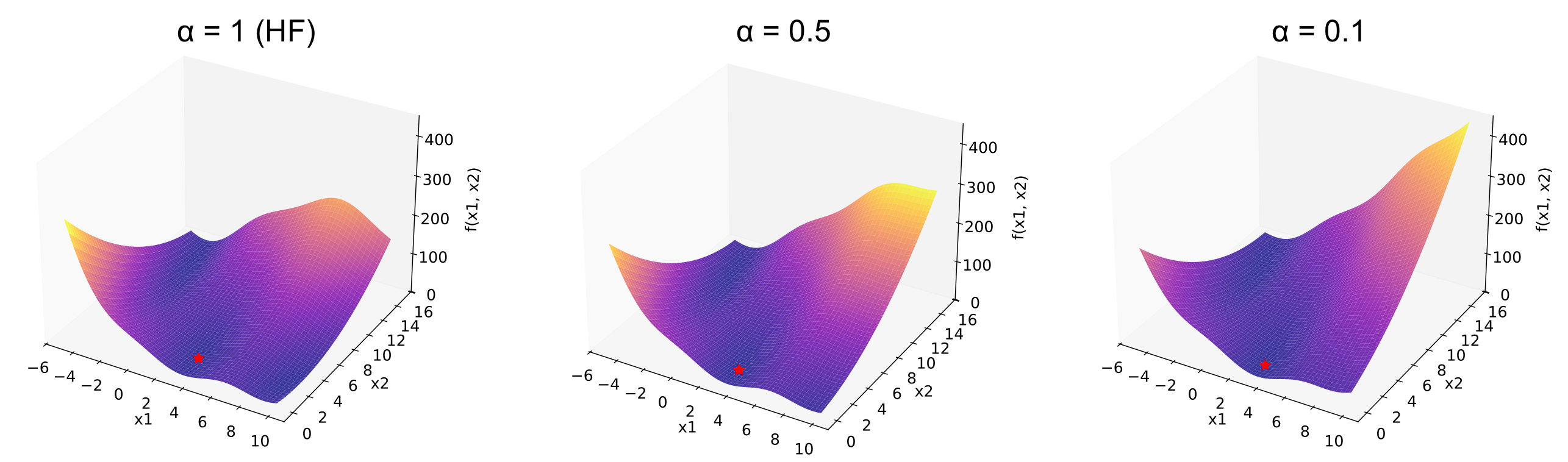}
    \caption{Branin surface with $\alpha$=1 (HF function), 0.5, and 0.1. Red stars show the location of one of the absolute minima in each surface.}
    \label{fig:branin_alpha}
\end{figure}

Figure \ref{fig:branin_alpha} illustrates the impact of varying $\alpha$ on the Branin function. When $\alpha$ = 1, the function retains its high-fidelity form. As $\alpha$ decreases, the function undergoes progressive deformation, particularly in the region $[4, 10] \times [6, 15]$, where both the curvature and overall topology of the surface are significantly altered.

Notably, the absolute minima of the function shift as $\alpha$ decreases. One of the three global minima is highlighted in the figure, showing a displacement trend with decreasing $\alpha$ but locating the minimum in the same region. This movement suggests that lower-fidelity versions of the function present structural variations while maintaining some correlation with the original HF surface. Although the function's absolute output values differ across fidelities, the lower-fidelity functions preserve key topological features, making them useful for approximating relative trends in optimization tasks. However, as mentioned in the main discussion, the interplay between cost and information of the LF function may create situations where the inclusion of a lower-fidelity source may be detrimental to the HF source optimization.

\textbf{Park-4D}
\begin{equation}
\begin{split}
f(\mathbf{x}, \alpha) &= \frac{x_1}{2} \left( \sqrt{1 + \frac{(x_2 + x_3^2)x_4}{x_1^2}} - 1\right) \\ & +  \left(x_1 + \left(3 - 1.5\left(1 - \alpha\right)\right)x_4\right) \exp(1 + \sin{x_3})
\end{split}
\end{equation}
defined over $[0, 1]^{4}$, and $\alpha \in [0, 1]$ is the bias term.

\subsection{Simple regret algorithm}

Algorithm for the standardized simple regret computation. Single-fidelity and multi-fidelity cumulative cost values ($c^{\rm {sf}}$ and $c^{\rm {mf}} $) represent the accumulated cost at each step of the optimization run.  
\begin{algorithm2e}[htbp]
    \SetAlgoLined
    \DontPrintSemicolon
    \SetNoFillComment
    \SetSideCommentLeft
    \KwIn{
        $t \in \mathbb{N_{+}}$ \tcp*{Number of single-fidelity optimization steps} \\
        $n \in \mathbb{N_{+}}$ \tcp*{Total number of multi-fidelity steps} \\
        $y^{\rm mf} \in \mathbb{R}^n$ \tcp*{Multi-fidelity output values}\\
        $l \in \{0,1\}^n$ \tcp*{Fidelity values (1 : high-fidelity, 0 : low-fidelity)}\\
        $c^{\rm {sf}} \in \mathbb{R}^t$ \tcp*{Single-fidelity cumulative cost values}\\
        $c^{\rm {mf}} \in \mathbb{R}^n$ \tcp*{Multi-fidelity cumulative cost values}\\
        $f^* \in \mathbb{R}$ \tcp*{High-fidelity global optimum}\\
    }
    \KwOut{$r \in \mathbb{R}^t$ \tcp*{Multi-fidelity regret over $t$ single-fidelity steps}}
    \BlankLine
    \SetKwFunction{FMain}{ComputeSimpleRegret}
    \SetKwProg{Fn}{Function}{:}{\KwRet ${r}$}
    \Fn{\FMain{$y^{\rm mf},\,{l},\,c^{\rm sf},\, c^{\rm mf}, \,f^*,\, t$}}{
        \tcp{Select high-fidelity outputs}
        ${y}^{\rm hf} \gets [y_{i}^{\rm mf}]_{\forall i | l_{i}=1}$\;
        \tcp{Compute high-fidelity simple regret}
        $r^{\rm hf} \gets f^* - y^{\rm hf}$\;
        \tcp{Initialize simple regret list}
        ${r} \gets [\,]$\;
        \For{$i \gets 1$ \KwTo $t$}{
            \tcp{Find minimum regret among indices where $c^{\rm mf}_{k} \leq c^{\rm sf}_{i}$}
            $r_{\rm min} \gets \min\left( [r_{k}^{\rm hf}]_{\forall k | c^{\rm mf}_{k} \leq c^{\rm sf}_{i} }\right)$\;
            \tcp{Append to simple regret list}
            Append $r_{\rm {min}}$ to ${r}$\;
        }
    }
    \caption{Algorithm for computing multi-fidelity simple regret}
    \label{alg:simple_regret}
\label{regret algo}
\end{algorithm2e}

\newpage
\subsection{Discount diagrams}\label{si: discount3d}
\begin{figure}
    \centering
    \includegraphics[width=0.98\linewidth]{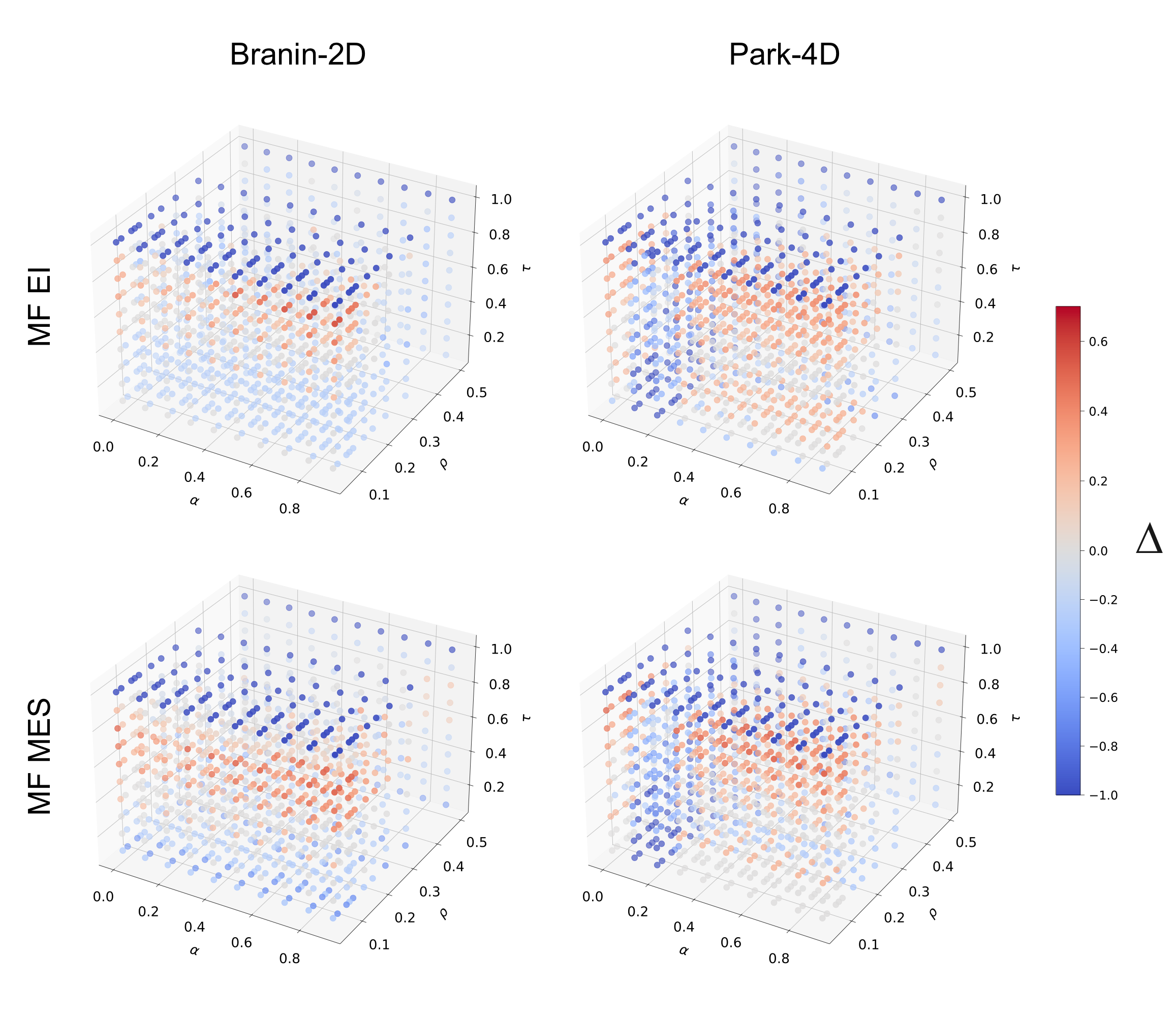}
    \caption{Informativeness of Branin-2D LF approximation varying with the $\alpha$ parameter}
    \label{3dplot}
\end{figure}

Figure \ref{3dplot} shows 3D-heatmaps of $\Delta$ as a function of $\rho$, $\alpha$ and $\tau$ for each synthetic function in the benchmark.

\clearpage
\subsection{Informativeness measurement}\label{R2problems}
The informativeness of the different problems is computed using the $R^2$ as explained in Metrics\ref{sec:metrics}.

\newpage
\begin{figure}
    \centering
    \includegraphics[width=0.98\linewidth]{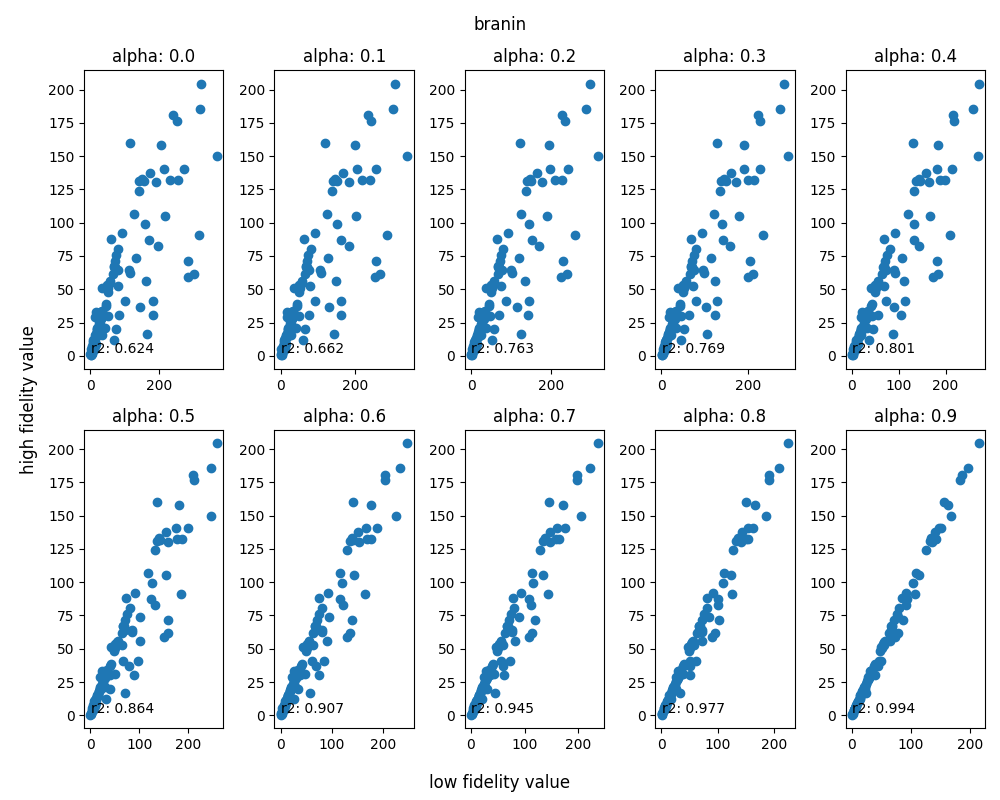}
    \caption{Informativeness of Branin-2D LF approximation varying with the $\alpha$ parameter}
    \label{braninr2}
\end{figure}

\begin{figure}
    \centering
    \includegraphics[width=0.98\linewidth]{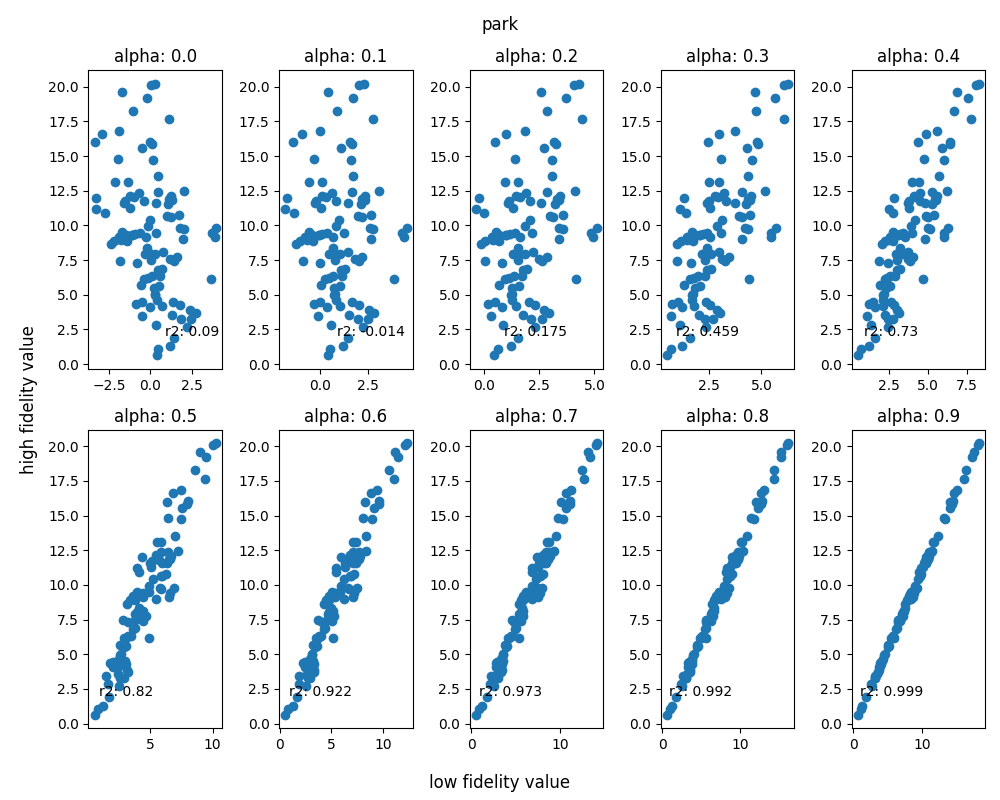}
    \caption{Informativeness of Park-2D LF approximation varying with the $\alpha$ parameter}
    \label{parkr2}
\end{figure}

\begin{figure}
    \centering
    \includegraphics[width=0.98\linewidth]{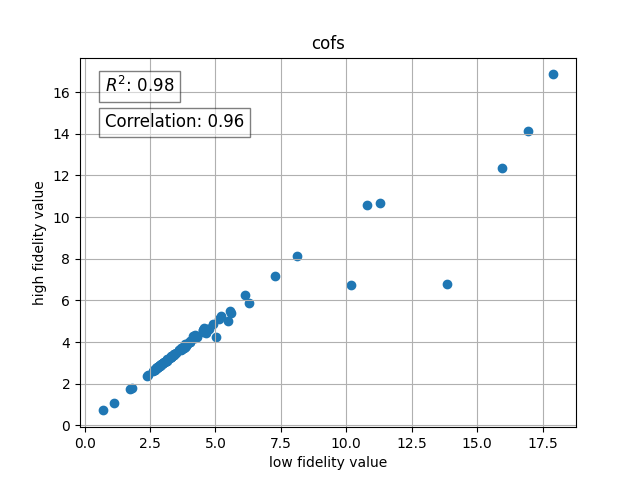}
    \caption{Informativeness of COFs LF approximation}
    \label{cofsr2}
\end{figure}

\begin{figure}
    \centering
    \includegraphics[width=0.98\linewidth]{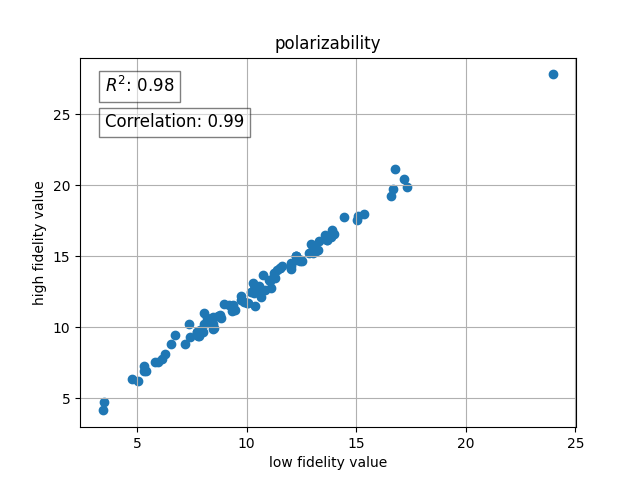}
    \caption{Informativeness of polarizability LF approximation}
    \label{polarizability}
\end{figure}

\begin{figure}[ht]
    \centering
    \includegraphics[width=0.98\linewidth]{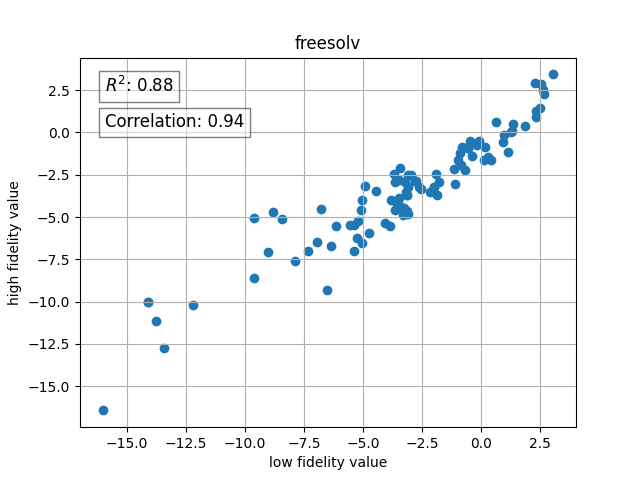}
    \caption{Informativeness of solvation energy LF approximation}
    \label{cofsr2}
\end{figure}

\clearpage

\subsection{Model extension to more than 3 fidelity levels in a homogeneous catalysis problem}\label{catalysis_task}

Here we show how the MF model can be extended to more than two fidelities in chemistry-relevant tasks. We focus on a high-throughput dataset of cross-coupling Pd-catalyzed reactions and build a three-fidelity regression problem to exemplify the adaptability of the technique. The dataset consists of 3955 reactions with their corresponding yields (optimization target), which were obtained from the combination of 3 bases, 4 ligands, 22 additives and 15 aryl halides\cite{ahneman2018predicting}.

% How we build the regressor
To simulate a task where more than two fidelities are available, we took the dataset and we split it in three subsets (5\% of data, 30\% and the rest). We featurized all the reactions as a 256-D vector using rxnfp\cite{schwaller2021mapping}, trained two Random Forest regressors on the low (5\% of the data) and medium (30\% of the data) data splits, and then predicted on the rest of the data using each trained model. The idea is building a low-fidelity predictor and a medium-fidelity (MedF) predictor that simulate cheaper approximations of the true yield (HF yield, the rest of the data). Figure \ref{BH_regression} shows the performance of the regressors in predicting the HF yield. The low-fidelity regressor gives a weak prediction performance ($R^{2}$ = 0.43) that is still able to inform about the true yield, whereas the medium-fidelity regressor gives a better performance ($R^{2}$ = 0.60). The dataset of 2592 reactions with the LF, MedF and HF yields is stored and used in a subsequent regression task.

% How we use the data
We then perform the regression task to emulate a MFBO scenario in 3 fidelities. We compare the performance of a standard SF GP using a Matern kernel and the MF GP used in this work. We select 10 evenly distributed samples for testing the performance of both models. For training, we use increasing random samples for the SF GP (10, 20, 30 and 40 samples respectively). For the MF GP, we use the same number of samples as the SF but assign 60\% to the HF, 30\% to the MedF assuming a cheaper cost of 0.1 (that is the corresponding amount of points divided by 0.1), and 10\% to the LF assuming a cost of 0.01. This sampling distribution emulates an MFBO scenario where the total sampling budget can be distributed among all the fidelities taking their cost into account. The kernel values are assigned as 0.8, 0.6 and 0.2 to the HF, MedF and LF respectively. Figure \ref{BH_regression} shows the prediction performance of the surrogates in this scenario of increasing sample sizes. The MF model consistently gets lower RMSE values and also lower uncertainties than the SF model. The higher MF performance illustrates how the model effectively leverages the information coming from the medium- and the low-fidelity regressors to improve the prediction error. This prediction capability can be therefore transferred to an MFBO campaign in a setup similar to that used in the binary setting in the study, and therefore reduce the overall optimization cost in this catalysis task.

\begin{figure}
    \centering
    \includegraphics[width=0.98\linewidth]{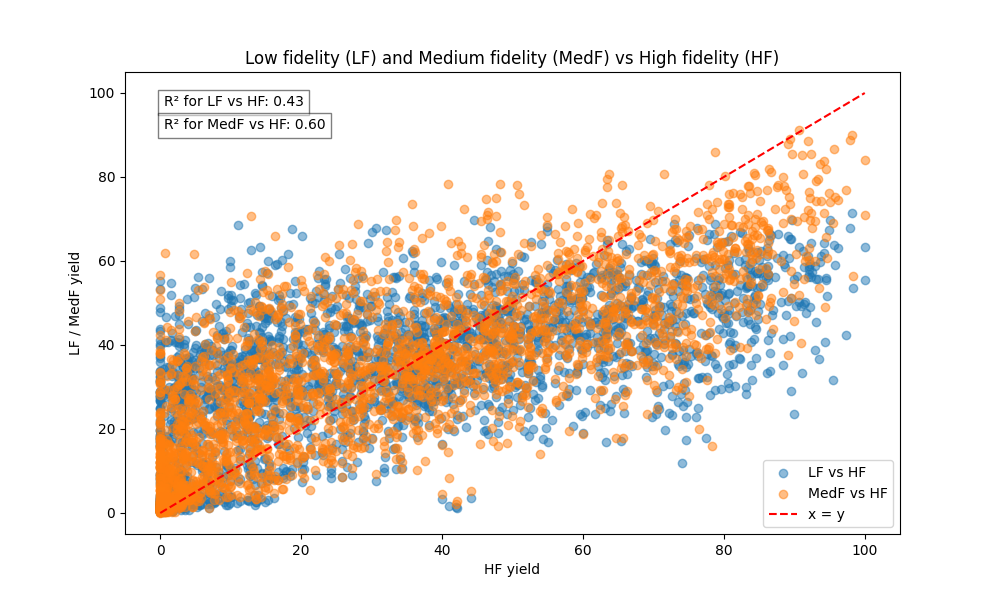}
    \caption{Low-fidelity regressor and medium-fidelity regressor performance on yield prediction for the Pd-catalyzed reaction task. MF: multi-fidelity, MedF: medium-fidelity, LF: low-fidelity.}
    \label{BH_dataset}
\end{figure}

\begin{figure}
    \centering
    \includegraphics[width=0.98\linewidth]{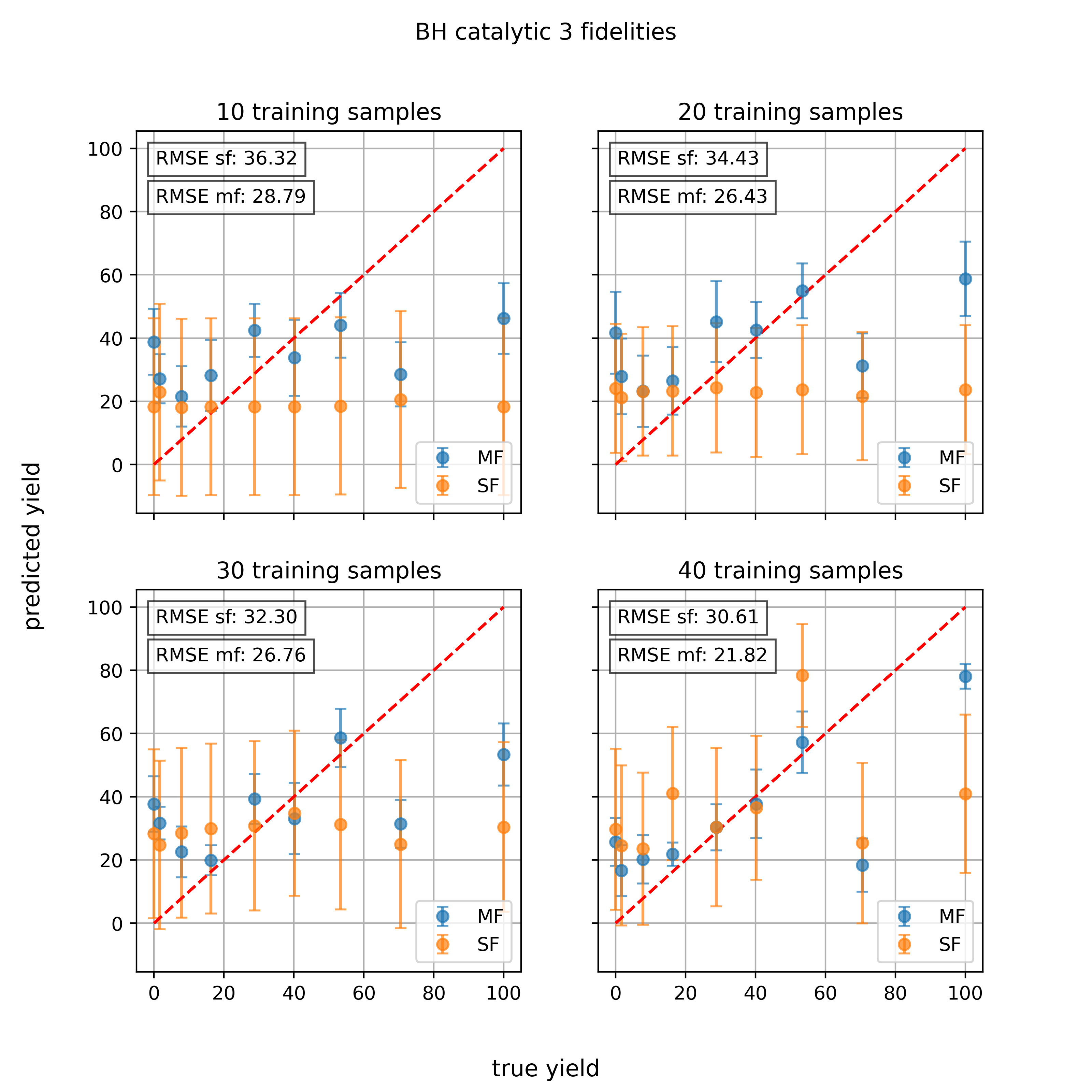}
    \caption{Regression performance of SF vs MF models in a catalysis task.}
    \label{BH_regression}
\end{figure}

\end{appendices}
\end{document}